\documentclass[10pt, a4paper]{article}
\usepackage[a4paper, margin=2.0cm]{geometry}
\usepackage[utf8]{inputenc}
\usepackage[T1]{fontenc}
\usepackage[english]{babel}
\usepackage{lmodern}
\usepackage{tikz}
\usetikzlibrary{arrows.meta, positioning, calc, backgrounds, fit, shapes.geometric}
\tikzset{needle/.style={circle, draw=blue!80!black, fill=blue!18, thick, minimum size=6.2mm, inner sep=0pt, text=blue!90!black, font=\sffamily\scriptsize\bfseries}}
\usepackage{rotating}
\usepackage{booktabs}
\usepackage{threeparttable}
\usepackage{graphicx}
\usepackage{amsmath, amssymb}
\usepackage{algorithm}
\usepackage{algpseudocode}
\usepackage{tikz}
\usetikzlibrary{arrows.meta, positioning, fit, backgrounds, calc}
\usepackage{multirow}
\usepackage{amssymb}
\usepackage{natbib}
\usepackage{soul}

\tikzset{keyleaf/.style={needle, draw=blue!70!black, fill=blue!10, text=blue!80!black}}

\usepackage{amsmath, amssymb, amsthm}
\usepackage{booktabs}
\usepackage{array}
\usepackage{tabularx}
\usepackage[x11names]{xcolor}
\usepackage{hyperref}

\usepackage{booktabs}
\usepackage{pifont}

\definecolor{treeblue}{RGB}{0, 102, 255}
\definecolor{treelight}{RGB}{229, 240, 255}
\definecolor{decgreen}{RGB}{68, 214, 0}
\definecolor{declight}{RGB}{235, 252, 227}
\definecolor{memorange}{RGB}{255, 109, 0}

\definecolor{PrimaryBlue}{RGB}{0, 102, 255}
\definecolor{AccentTeal}{RGB}{0, 191, 165}
\definecolor{AccentAmber}{RGB}{255, 145, 0}
\definecolor{AccentPurple}{RGB}{138, 0, 230}
\definecolor{SoftBlue}{RGB}{229, 240, 255}
\definecolor{SoftTeal}{RGB}{224, 248, 244}
\definecolor{SoftAmber}{RGB}{255, 244, 224}
\definecolor{SoftPurple}{RGB}{243, 229, 255}
\definecolor{SlateGray}{RGB}{71, 85, 105}
\definecolor{BorderGray}{RGB}{203, 213, 225}

\hypersetup{
    colorlinks=true,
    linkcolor=PrimaryBlue,
    citecolor=PrimaryBlue,
    urlcolor=PrimaryBlue
}

\definecolor{limegreen}{HTML}{44D600}

\definecolor{memlight}{RGB}{255, 240, 224}
\definecolor{gpupurple}{RGB}{138, 0, 230}
\definecolor{gpulight}{RGB}{243, 229, 255}
\definecolor{ncclred}{RGB}{245, 0, 87}
\definecolor{nccllight}{RGB}{254, 229, 238}
\definecolor{panelbg}{RGB}{248, 249, 251}
\definecolor{panelborder}{RGB}{200, 206, 214}
\definecolor{embgray}{RGB}{236, 239, 244}

\definecolor{leafblue}{HTML}{0066FF}
\definecolor{leafbg}{HTML}{E5F0FF}
\definecolor{bindteal}{HTML}{00BFA5}
\definecolor{bindbg}{HTML}{E0F8F4}
\definecolor{treeindigo}{HTML}{5500FF}
\definecolor{treebg}{HTML}{EAE5FF}
\definecolor{slotpurple}{HTML}{8A00E6}
\definecolor{slotbg}{HTML}{F3E5FF}
\definecolor{rootcrimson}{HTML}{F50057}
\definecolor{rootbg}{HTML}{FEE5EE}
\definecolor{tailamber}{HTML}{FF9100}
\definecolor{tailbg}{HTML}{FFF4E0}
\definecolor{quadgreen}{HTML}{44D600}
\definecolor{skiporange}{HTML}{FF6D00}
\definecolor{decslate}{RGB}{51, 65, 85}
\definecolor{decbg}{HTML}{F1F5F9}

\tikzset{
    filler/.style={circle, draw=gray!60, fill=gray!12, minimum size=6.2mm, inner sep=0pt, text=gray!80},
    needle/.style={circle, draw=blue!80!black, fill=blue!18, thick, minimum size=6.2mm, inner sep=0pt, text=blue!90!black, font=\sffamily\scriptsize\bfseries},
    treenode/.style={circle, draw=teal!80!black, fill=teal!12, thick, minimum size=6.5mm, inner sep=0pt, text=teal!90!black, font=\sffamily\scriptsize\bfseries},
    rootnode/.style={circle, draw=teal!90!black, fill=teal!28, very thick, minimum size=7.8mm, inner sep=0pt, font=\sffamily\scriptsize\bfseries},
    poolnode/.style={rectangle, rounded corners=3pt, draw=purple!80!black, fill=purple!12, thick, minimum width=28mm, minimum height=6.5mm, font=\sffamily\scriptsize\bfseries},
    shift_edge/.style={-{Stealth[scale=0.75]}, draw=orange!85!black, thick, bend bottom=30},
    active_tree/.style={-{Stealth[scale=0.8]}, draw=blue!75!black, very thick},
    muted_tree/.style={-{Stealth[scale=0.7]}, draw=gray!45, dashed, thin},
    attn_mesh/.style={draw=red!35, ultra thin, opacity=0.55},
    attn_needle/.style={draw=red!75!black, thin, opacity=0.75}
}

\hypersetup{
    colorlinks=true,
    linkcolor=blue!60!black,
    citecolor=blue!60!black,
    urlcolor=blue!60!black
}

\title{\textcolor{limegreen}{LimeTree:} Linear-Memory Graph Neural Networks over Tree-Graphs for Resource-Efficient Language Modeling and Context Extrapolation}
\author{
  \textbf{Elisabeth Fink} \\
  \footnotesize Middlesex University, Department of Computer Science\\ \footnotesize The Burroughs, Hendon, London NW4 4BT, UK.\\ \footnotesize \textbf{e.fink@mdx.ac.uk}}
  
\date{\today}

\begin{document}

\maketitle

\begin{figure}[htbp]
    \centering
    \includegraphics[width=0.5\linewidth]{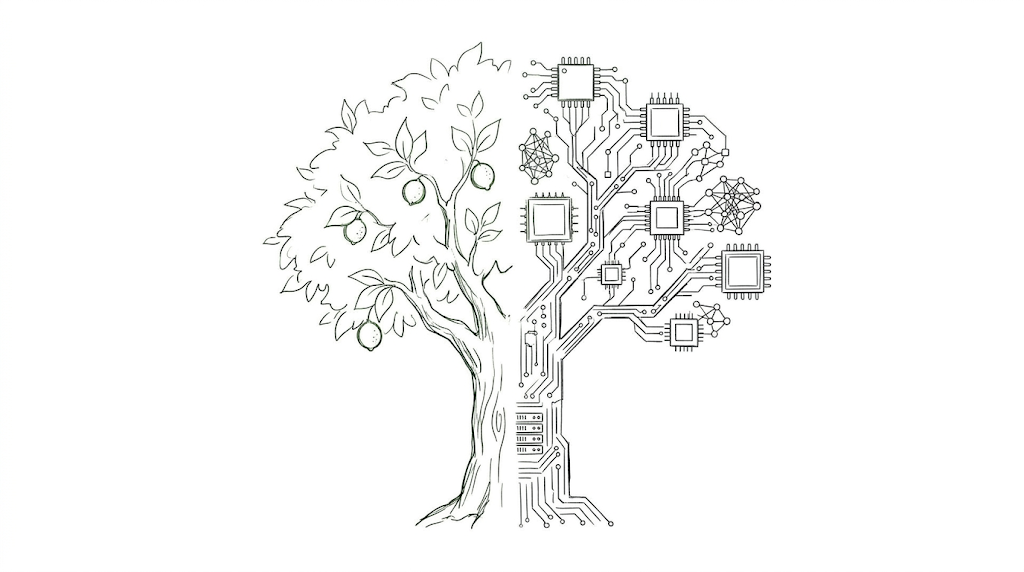}
\end{figure}

\begin{abstract}
The search for computationally efficient alternatives to the Transformer has yielded numerous architectures in recent years; however, most exhibit severe weaknesses in associative recall—such as key-value (KV) retrieval—and length extrapolation, where even standard Transformers struggle. This paper introduces LimeTree, a linear-complexity architecture in both memory and compute for sequence modeling. By integrating synthetic KV retrieval tasks directly into standard language modeling training, LimeTree achieves strong perplexity alongside reliable recall. Two variants were evaluated on WikiText-103 and PG-19 under strict compute constraints. On WikiText-103, LimeTree reaches a token-level perplexity of 19 in just 6 hours on four Quadro RTX 8000 GPUs, while maintaining 100\% KV retrieval accuracy even when extrapolating from a training context length of 512 to an evaluation length of 16,384. On PG-19, the model achieves a token-level perplexity of 10.84 in 35.5 hours on a single RTX PRO 6000 Blackwell GPU with minimal hyperparameter tuning. These results position LimeTree as a competitive linear-complexity alternative to Transformers, combining superior perplexity and training throughput with robust in-context retrieval and length extrapolation.
\end{abstract}

\section*{Highlights}
\begin{itemize}
    
\item Predictable, Linear Infrastructure Costs: Standard AI models (Transformers) become attention cost grows quadratically with input length as inputs get longer, driving up cloud compute and hardware bills. LimeTree’s memory and compute scale strictly linearly, enabling lightweight models that process massive contexts at a fraction of standard deployment and inference costs.

\item Train Cheaply on Short Data, Deploy on Massive Contexts: Models typically require exorbitant budgets to train on long documents. LimeTree can be trained affordably on short sequences (4,096 tokens) and immediately process 128× longer inputs (over 524,000 tokens—roughly 1,000 pages of text) out of the box with perplexity rising only from 10.84 to 11.38 (key-value recall is 100\% at most lengths up to 128k, 81\% at 256k, 50\% at 512k).

\item Proof of concept: token-level perplexity in the range of published token-level baselines, trained on a single GPU. Word-level PG-19 perplexity (37.4) is behind Transformer-XL (36.3) and later models.

\end{itemize}

\section{Introduction}

Since its introduction, the Transformer architecture as presented in \cite{vaswani2017attention} has become the foundational backbone of modern artificial intelligence, evolving from early sequence-to-sequence machine translation models into general-purpose foundation models across natural language processing, computer vision, and multimodal domains. More recently, the capabilities of these models have been substantially expanded by augmenting next-token prediction with test-time compute \cite{snell2024scaling}, explicit reasoning chains \cite{wei2022chain}, and tree-search algorithms such as Monte Carlo Tree Search \cite{kocsis2006bandit,silver2016mastering,yao2023tree,hao2023reasoning}. By integrating these search mechanisms, language models have transitioned from passive text generators into autonomous agents capable of complex planning, iterative self-correction, and multi-step problem solving. However, this shift toward deep reasoning and expansive search trees drastically
increases the number of generated tokens and active context requirements,
amplifying the underlying computational bottlenecks of the standard
self-attention mechanism \citep{vaswani2017attention}. Chain-of-thought
prompting \citep{wei2022chain}, tree-structured search \citep{yao2023tree},
scaling of test-time compute \citep{snell2024scaling}, and reinforcement-learned
reasoning models \citep{openai2024o1,deepseek2025r1} all trade longer
generations for accuracy.

To mitigate these scaling costs, extensive research has focused on developing
sub-quadratic alternatives to attention. Early linear-attention variants, such as
Linformer \citep{wang2020linformer} and Performer \citep{choromanski2021rethinking},
as well as pure recurrent formulations like Receptance Weighted Key Value
architectures \citep{peng2023rwkv}, demonstrated theoretical efficiency gains but
have seen limited adoption in publicly documented frontier systems, with
performance degradation relative to softmax attention reported in recall-sensitive
settings \citep{katharopoulos2020transformers,arora2024zoology}. Similarly, early
structured state-space models (SSMs) like S4 \citep{gu2022efficiently},
H3 \citep{fu2023hungry}, and Hyena \citep{poli2023hyena} largely remain confined
to specialized research settings or domain-specific tasks. In contrast, selective
state-space models---most notably Mamba \citep{gu2024mamba} and
Mamba-2 \citep{dao2024transformers}---alongside gated linear RNNs such as
Griffin \citep{de2024griffin}, xLSTM \citep{beck2024xlstm}, and
RetNet \citep{sun2023retentive}, have achieved meaningful practical traction.
Even among these stronger contenders, pure sub-quadratic architectures exhibit
well-documented weaknesses in copying and in-context recall
\citep{jelassi2024repeat,arora2024zoology,waleffe2024empirical}; consequently,
among openly documented production-scale systems, hybrid architectures-such as
Jamba \citep{lieber2024jamba}, Zamba \citep{glorioso2024zamba},
MiniMax-01 \citep{minimax2025minimax01}, and Nemotron-H \citep{nvidia2025nemotronh}---predominate,
interleaving linear layers with full-attention layers to preserve high-resolution
recall.

While Transformers prevail in this space, models trained on a fixed maximum length typically degrade sharply on longer inputs~\citep{deletang2023neural}—a failure strongly tied to positional encoding. Both sinusoidal and learned absolute embeddings~\citep{vaswani2017attention}, as well as unmodified rotary embeddings~\citep{su2024roformer,barbero2025round}, extrapolate poorly beyond the training length. To improve extrapolation, Attention with Linear Biases (ALiBi)~\citep{press2022train} replaces positional embeddings with a fixed, distance-proportional penalty on attention logits, followed by relative-position schemes like xPos~\citep{sun2023length} and FIRE~\citep{li2024functional}, or randomized positional encodings~\citep{ruoss2023randomized}. Remarkably, controlled comparisons on algorithmic and reasoning tasks show that decoder-only Transformers without explicit positional encodings can out-generalize ALiBi, rotary, and T5-style relative biases~\citep{kazemnejad2023impact}. For already-pretrained models, context windows can be extended by rescaling rotary frequencies, as in Position Interpolation~\citep{chen2023extending} and YaRN~\citep{peng2024yarn}, whereas StreamingLLM~\citep{xiao2024efficient} stabilizes infinite-stream generation via attention sinks and a sliding window (albeit discarding past context). Theoretically, length generalization has been characterized through formal frameworks with provable conditions~\citep{huang2025formal}, linked to whether a task admits a short program in a restricted language~\citep{zhou2024what}, and attributed in part to softmax attention losing sharpness as token count grows~\citep{velickovic2024softmax}.

Beyond these extrapolation limits, standard Transformers are also bottlenecked by quadratic $\mathcal{O}(N^2)$ compute and linear $\mathcal{O}(N)$ KV-cache memory growth. Yet replacing full attention with sub-quadratic alternatives exposes a fundamental tension in sequence modeling: compressing arbitrarily long histories into a fixed-size recurrent state creates an information bottleneck that degrades exact in-context retrieval, associative recall (such as key-value lookup), and zero-shot length extrapolation. Recent analyses~\citep{Fichtl_2026} argue that these structural deficiencies in precise memory routing prevent linear-complexity models from matching the algorithmic fidelity of full attention. Graph Neural Networks (GNNs)~\citep{scarselli2009graph,kipf2017semi,gilmer2017neural}, which propagate information across structured relational topologies via iterative neighborhood aggregation, offer a compelling alternative and have recently demonstrated strong length extrapolation on key-value retrieval tasks~\citep{fink2026gtformer}. However, while GNNs have been applied in natural language processing to encode syntactic dependency trees or word-document co-occurrence graphs~\citep{yao2019graph,wu2023graph}, and to augment pretrained language models with corpus-level retrieval graphs~\citep{meng2022gnnlm}, their potential as a standalone, syntax-free causal backbone for sequence modeling remains largely unexplored.

To resolve these limitations without sacrificing efficiency, this paper introduces LimeTree (\textcolor{limegreen}{li}near-\textcolor{limegreen}{me}mory graph-\textcolor{limegreen}{tree}), a purely linear-complexity architecture that reconciles language modeling quality with exact associative recall. Expanding upon the graph-tree formulation \cite{fink2026gtformer}, LimeTree adapts the topology with novel architectural features tailored for autoregressive token-sequence modeling. Specifically, LimeTree reformulates sequence modeling as a graph problem by constructing a hierarchical tree directly over the input tokens and applying GNN message passing across the resulting edges. Rather than trading retrieval fidelity for sub-quadratic scaling, the training regime integrates key-value lookup objectives directly alongside language modeling, enabling efficient long-range routing and exact associative recall within a strictly linear compute and memory footprint. Empirical evaluations show that LimeTree scales reliably across multi-GPU configurations to large corpora, achieving a competitive token-level perplexity of $19.46$ on WikiText-103 and $10.84$ (corresponding to a word-level perplexity of $37.49$) on the long-context PG-19 benchmark, while maintaining $100\%$ accuracy on synthetic key-value retrieval tasks. Crucially, the architecture exhibits exceptional zero-shot length extrapolation: it preserves close to $100\%$ retrieval accuracy on sequences $32\times$ to $128\times$ longer than its training context, and when trained on sequences of at most $4,096$ tokens, maintains a stable perplexity below $12$ ($11.38$) out to $524,288$ tokens—the hardware memory limit of a single Nvidia Quadro RTX 6000 Pro BSE GPU—thereby overcoming the primary bottlenecks that have historically restricted standalone sub-quadratic models.

\section{Methods}

First, we introduce the two benchmark datasets Wikitext 103 and PG19 as well as the setup for the added key-value pair task. This is followed by the construction of the graph-tree over the token sequence and the architecture of the GNN, which is slightly different for models on the two datasets. The model on Wikitext is somewhat simpler, reflecting the smaller size and higher perplexity scores that are possible to achieve. The model for PG19 has some added features. At the end of this section we outline how the setup is adapted to work on multiple GPUs.

\subsection{Datasets and Evaluation}

\textbf{Word-to-Token:} Text is converted into tokens, each of which is a short word or a subword fragment. We use the \texttt{GPT2TokenizerFast} implementation from the Hugging Face \texttt{transformers} library~\cite{wolf-etal-2020-transformers}, which is backed by the Rust-based \texttt{tokenizers} library~\cite{moi2023huggingface}. The tokenizer applies byte-level Byte-Pair Encoding (BPE) with a vocabulary of 50,257 tokens. Because the input is first mapped to bytes, any Unicode string can be encoded without out-of-vocabulary tokens; a leading space is encoded as part of the following token and is displayed as the special character \texttt{\.{G}}. The fast implementation produces the same tokenization as the pure-Python GPT-2 tokenizer while being substantially faster on large batches, and it additionally provides offset mappings that link each token to its character span in the original string.

\textbf{WikiText-103~\cite{merity2018analysis}:} WikiText-103 is a word-level language modeling benchmark released by Salesforce Research. It comprises over 103 million training tokens drawn from 28{,}475 ``Good'' and ``Featured'' Wikipedia articles, with a vocabulary of roughly 267{,}000 word types (all words occurring at least three times in the training set; remaining words are mapped to \texttt{<unk>}). The original casing, punctuation, and numbers are preserved. Because the corpus consists of full-length articles rather than shuffled sentences, it is well suited to evaluating a model's ability to exploit long-range context. 

\textbf{PG-19~\cite{rae2019compressivetransformerslongrangesequence}:} PG-19 is a long-form language modeling benchmark introduced by researchers at DeepMind, consisting of books from Project Gutenberg published before 1919. It contains 28{,}752 books, split into train (1{,}973{,}136{,}207 tokens), validation (3{,}007{,}061 tokens), and test (6{,}966,499 tokens) sets. The average book is about 69k words long, so individual documents span tens to hundreds of thousands of tokens, making the dataset a natural testbed for long-range dependencies. Its relatively small test set makes evaluation inexpensive. PG-19 is widely used to assess long-context perplexity in context-extension methods such as Position Interpolation \cite{chen2023extending}, YaRN \cite{peng2024yarn}, LongLoRA \cite{chen2024longlora}, and StreamingLLM \cite{xiao2024efficient}. In the typical protocol, a pretrained model has its context window extended and perplexity is then measured as a function of context length on the test set or a subset of its books.

\textbf{Key-value (KV) retrieval task data:} In-context associative recall—recovering the most recent value bound to a queried key across long sequences—is a foundational capability for language modeling that typically breaks down in sub-quadratic architectures due to fixed-state memory bottlenecks~\citep{olsson2022incontext,fu2023hungry,arora2024zoology,fink2026gtformer}. To jointly enforce and evaluate exact long-range memory routing alongside natural language modeling, training and evaluation batches are augmented with a synthetic, unaligned KV retrieval task drawn from a disjoint symbol vocabulary offset beyond the $50{,}257$ GPT-2 tokens ($200$ background filler tokens, $3$ distinct keys $\{k_0, k_1, k_2\}$, and $4$ discrete values $\{v_0, \dots, v_3\}$). Each sequence of length $N$ is initialized with uniform random filler tokens and seeded with $\max(2, \lfloor N / 16 \rfloor)$ unassociated distractor values at even positions to prevent bag-of-words shortcuts. The context is partitioned into six equal contiguous segments organized into two sequential waves of three segments each; within each wave, a random permutation of the three keys is assigned freshly sampled values at odd-aligned token pairs $(2m+1, 2m+2)$. This odd alignment deliberately forces every key--value binding to straddle the binary tree's even--odd leaf merge boundaries, testing cross-branch information routing rather than local pooling. Because the second wave overwrites all three keys from the first wave, the model must perform both cross-boundary token binding and temporal state tracking across the full context before autoregressively predicting the final values of $(k_0, k_1, k_2)$, evaluated via joint Exact-3 sequence accuracy.

\textbf{Evaluation with the perplexity (PPL) score:} For a token sequence $x = (x_1,\dots,x_N)$ and a model $p_\theta$ that factorises autoregressively,
$$
\mathrm{PPL}(x) \;=\; \exp\!\Big(-\frac{1}{N}\sum_{i=1}^{N}\log p_\theta(x_i \mid x_{<i})\Big) \;=\; p_\theta(x_1,\dots,x_N)^{-1/N},
$$ where $p_\theta(x_i \mid x_{<i})$ is the probability that the model assigns $x_i$ as the next token given the sequence up until $x_{i-1}$, or how confident it was with the next token. We have $\mathrm{PPL}\ge 1$, with equality only if the model assigns probability 1 to every observed token. A uniform model over a vocabulary of size $V$ has $\mathrm{PPL}=V$. This is the "effective branching factor" interpretation. Mathematically, a PPL of 20 means that, on average, the model is as uncertain as if it were choosing equally between 20 different possible tokens at every single step. With a context limit $L$, the conditional becomes $p_\theta(x_i \mid x_{i-k_i},\dots,x_{i-1})$, where $k_i\le L-1$ is the context actually available to token $i$.

\subsection{Token Sequence-to-Graph-Tree Construction 1 (Wikitext)}

Given a prefix sequence of $N$ context token indices $\mathbf{c} = (c_1, \dots, c_N)$, initial representations are obtained via scaled token embeddings with dropout:
\begin{equation}
    \mathbf{h}_i = \text{Dropout}\left(\mathbf{E}[c_i] \sqrt{d}\right) \in \mathbb{R}^d.
\end{equation}
To capture unaligned local token pairs before hierarchical reduction, a forward bigram leaf-binding module combines each normalized token $\bar{\mathbf{h}}_i = \text{LN}(\mathbf{h}_i)$ with its right neighbor $\bar{\mathbf{h}}_{i+1} = \text{LN}(\mathbf{h}_{i+1})$ (zero-padded at $i=N$) via bilinear interaction:
\begin{equation}
    \mathbf{z}_i = \text{GELU}(\mathbf{W}_L \bar{\mathbf{h}}_i) \odot (\mathbf{W}_R \bar{\mathbf{h}}_{i+1}), \quad
    \mathbf{h}^{(0)}_i = \mathbf{W}_{\text{skip}} \bar{\mathbf{h}}_i + \text{MLP}_{\text{bind}}\left([\bar{\mathbf{h}}_i \,\|\, \bar{\mathbf{h}}_{i+1} \,\|\, \mathbf{z}_i]\right).
\end{equation}
A direct global leaf summary $\mathbf{r}_{\text{direct}} = \mathbf{W}_{\text{root}}\text{LN}\left(\frac{1}{N}\sum_{i=1}^N \mathbf{h}^{(0)}_i\right)$ is computed in parallel to serve as a 1-hop residual bypass to the tree root.

Starting from leaf states $\mathbf{H}^{(0)} \in \mathbb{R}^{N \times d}$, the hierarchy is constructed iteratively across levels $\ell = 0, 1, \dots, L-1$ until a single root node remains ($N_L = 1$). At each level $\ell$ with sequence length $N_\ell > 1$:
\begin{enumerate}
    \item \textbf{Fenwick Tail Extraction:} The rightmost up to two nodes $\mathbf{H}^{(\ell)}_{N_\ell-1:N_\ell}$ are augmented with level embeddings $\mathbf{e}^{\text{lvl}}_{\min(\ell, 8)}$ and binary branch embeddings $\mathbf{e}^{\text{branch}}_{0,1}$, then appended to a multi-scale Fenwick tail list.
    \item \textbf{Bottom-Up Reduction:} Adjacent pairs are merged via the GNN tree cell to produce $\tilde{\mathbf{H}}^{(\ell+1)} = \text{BottomUp}(\mathbf{H}^{(\ell)}) \in \mathbb{R}^{\lceil N_\ell / 2 \rceil \times d}$.
    \item \textbf{Alternating 2-Hop Quad-Tree Shortcut:} On odd transitions where a grandparent level $\mathbf{H}^{(\ell-1)}$ is cached, every 4 grandchildren are pooled and injected into $\tilde{\mathbf{H}}^{(\ell+1)}$ via $\text{QuadShortcut}(\mathbf{H}^{(\ell-1)}, \tilde{\mathbf{H}}^{(\ell+1)})$, after which the cache is cleared.
    \item \textbf{Subtree Slot Capture:} The first intermediate level satisfying $N_{\ell+1} \le K_{\text{slots}}$ (with $K_{\text{slots}} = 32$) is recorded as the coarse subtree state $\mathbf{H}_{\text{mid}}$.
\end{enumerate}

Once the bottom-up pass terminates at $\mathbf{H}^{(L)} \in \mathbb{R}^{1 \times d}$, the root representation incorporates the 1-hop leaf bypass:
\begin{equation}
    \mathbf{h}_{\text{root}} = \mathbf{H}^{(L)} + \tanh(\alpha_{\text{root}}) \, \mathbf{r}_{\text{direct}}.
\end{equation}
The captured subtree nodes $\mathbf{H}_{\text{mid}} \in \mathbb{R}^{M \times d}$ ($M \le K_{\text{slots}}$) are then contextualized by $\mathbf{h}_{\text{root}}$ via top-down refinement, yielding refined subtree slots $\hat{\mathbf{H}}_{\text{mid}}$ and a secondary summary vector $\mathbf{h}_{\text{td}}$. A 1-hop adaptive average pooling bypass from the leaves $\mathbf{H}^{(0)}$ and learnable positional embeddings $\mathbf{P}_{1:M}$ are added to the refined slots:
\begin{equation}
    \mathbf{M}_{\text{mid}} = \hat{\mathbf{H}}_{\text{mid}} + \tanh(\alpha_{\text{slot}}) \, \mathbf{W}_{\text{slot}}\text{LN}\left(\text{AvgPool}_{M}(\mathbf{H}^{(0)})\right) + \mathbf{P}_{1:M}.
\end{equation}
Finally, the global root vector is fused with the top-down summary via $\mathbf{h}_{\text{final}} = \mathbf{h}_{\text{root}} + \mathbf{W}_{\text{merge}}[\mathbf{h}_{\text{root}} \,\|\, \mathbf{h}_{\text{td}}]$. The encoder outputs a normalized multi-scale memory bank $\mathbf{M} = \text{LN}([\mathbf{h}_{\text{final}} \,\|\, \mathbf{M}_{\text{mid}} \,\|\, \mathbf{T}_{-24:}])$—concatenating the root, refined subtree slots, and the last 24 Fenwick tail states—alongside a pooled root context vector $\mathbf{c}_{\text{ctx}} = \text{MLP}_{\text{ctx}}(\mathbf{h}_{\text{final}})$.

\subsection{Token Sequence-to-Graph-Tree Construction 2 (PG19)}
\label{subsec:seq_to_tree}

To achieve sub-quadratic long-context language modeling on PG-19 while preserving $100\%$ exact key-value recall across arbitrary token alignments, LimeTree transforms an input context sequence of length $L$ into a multi-scale graph-tree hierarchy and processes it via a specialized bidirectional Graph Neural Network (\texttt{GraphTreeCell}). Below, we detail the sequence-to-graph-tree construction pipeline and the message-passing architecture.

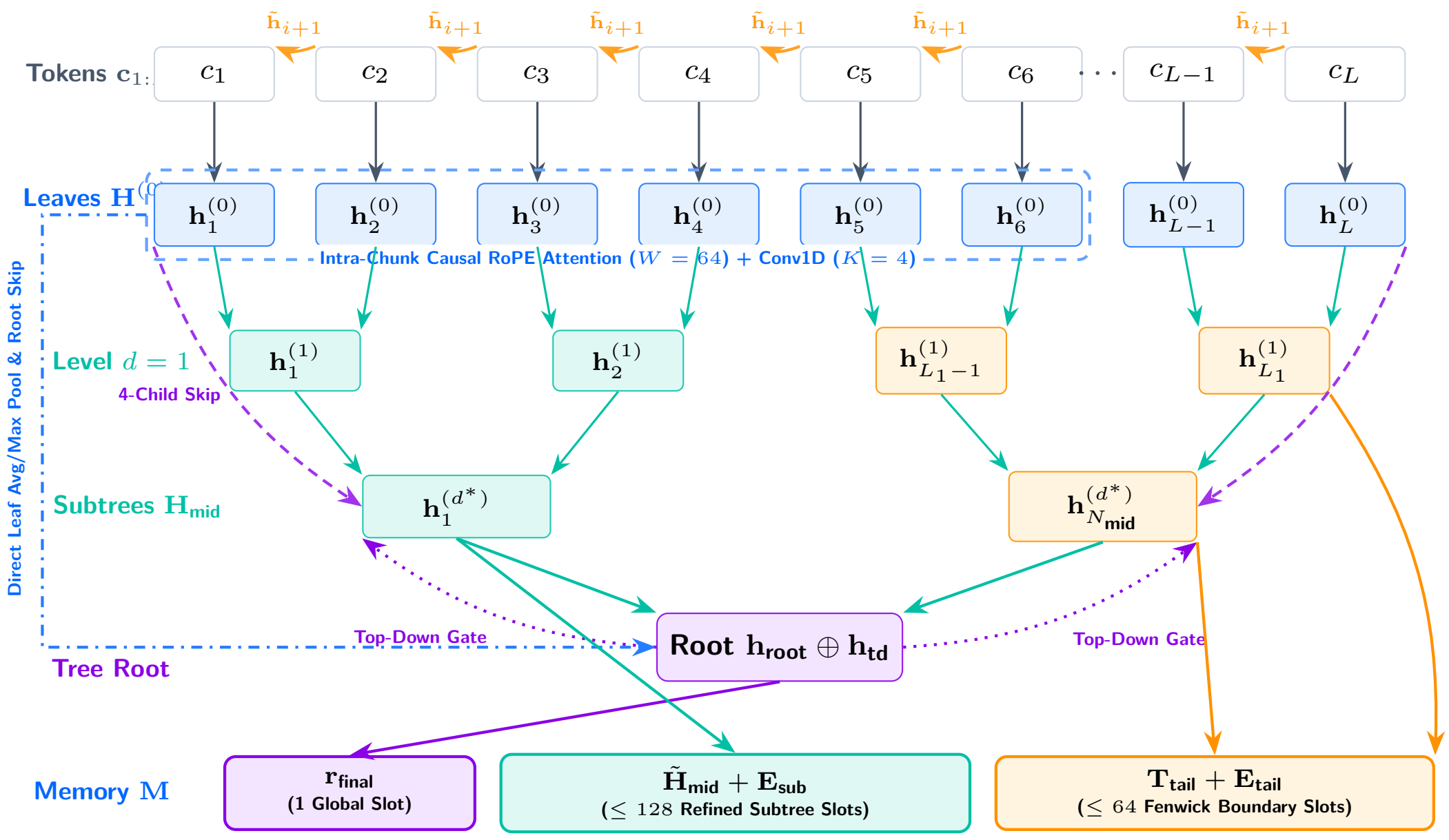
\begin{figure}[htbp]
\centering
\begin{tikzpicture}[
    font=\sffamily\footnotesize,
    >=Stealth,
    tok/.style={draw=BorderGray, fill=white, minimum width=1.15cm, minimum height=0.52cm, rounded corners=2pt, align=center},
    leaf/.style={draw=PrimaryBlue!80, fill=SoftBlue, minimum width=1.15cm, minimum height=0.6cm, rounded corners=2pt, align=center, font=\sffamily\scriptsize\bfseries},
    tree_node/.style={draw=AccentTeal!85, fill=SoftTeal, minimum width=1.25cm, minimum height=0.58cm, rounded corners=2pt, align=center, font=\sffamily\scriptsize\bfseries},
    tail_node/.style={draw=AccentAmber!90, fill=SoftAmber, minimum width=1.25cm, minimum height=0.58cm, rounded corners=2pt, align=center, font=\sffamily\scriptsize\bfseries},
    root_node/.style={draw=AccentPurple!90, fill=SoftPurple, minimum width=2.2cm, minimum height=0.65cm, rounded corners=3pt, align=center, font=\sffamily\footnotesize\bfseries},
    mem_box/.style={draw=PrimaryBlue!90, thick, fill=SoftBlue!60, rounded corners=3pt, minimum height=0.7cm, align=center, font=\sffamily\scriptsize\bfseries}
]

\node[anchor=east, font=\sffamily\scriptsize\bfseries, text=SlateGray] at (-0.3, 0) {Tokens $\mathbf{c}_{1:L}$};
\foreach \i/\lbl in {0/$c_1$, 1/$c_2$, 2/$c_3$, 3/$c_4$, 4/$c_5$, 5/$c_6$, 6/$c_{L-1}$, 7/$c_L$} {
    \node[tok] (t\i) at (\i*1.55, 0) {\lbl};
}

\foreach \i/\j in {0/1, 1/2, 2/3, 3/4, 4/5, 6/7} {
    \draw[->, AccentAmber!85, thick, bend left=22] (t\j.north west) to node[above, font=\sffamily\tiny, text=AccentAmber!95] {$\tilde{\mathbf{h}}_{i+1}$} (t\i.north east);
}
\node[font=\sffamily\bfseries, text=SlateGray] at (5.5*1.55 - 0.02, 0) {$\cdots$};

\node[anchor=east, font=\sffamily\scriptsize\bfseries, text=PrimaryBlue] at (-0.3, -1.15) {Leaves $\mathbf{H}^{(0)}$};
\foreach \i/\lbl in {0/$\mathbf{h}^{(0)}_1$, 1/$\mathbf{h}^{(0)}_2$, 2/$\mathbf{h}^{(0)}_3$, 3/$\mathbf{h}^{(0)}_4$, 4/$\mathbf{h}^{(0)}_5$, 5/$\mathbf{h}^{(0)}_6$, 6/$\mathbf{h}^{(0)}_{L-1}$, 7/$\mathbf{h}^{(0)}_L$} {
    \node[leaf] (l\i) at (\i*1.55, -1.35) {\lbl};
    \draw[->, SlateGray, semithick] (t\i.south) -- (l\i.north);
}

\draw[draw=PrimaryBlue!60, dashed, thick, rounded corners=3pt] (-0.65, -1.78) rectangle (5*1.55+0.65, -0.92);
\node[fill=white, inner sep=1.5pt, font=\sffamily\tiny\bfseries, text=PrimaryBlue] at (2.5*1.55, -1.78) {Intra-Chunk Causal RoPE Attention ($W=64$) + Conv1D ($K=4$)};

\node[anchor=east, font=\sffamily\scriptsize\bfseries, text=AccentTeal] at (-0.1, -2.75) {Level $d=1$};
\node[tree_node] (d1_0) at (0.775, -2.75) {$\mathbf{h}^{(1)}_1$};
\node[tree_node] (d1_1) at (3.875, -2.75) {$\mathbf{h}^{(1)}_2$};
\node[tail_node] (d1_2) at (6.975, -2.75) {$\mathbf{h}^{(1)}_{L_1-1}$};
\node[tail_node] (d1_3) at (10.075, -2.75) {$\mathbf{h}^{(1)}_{L_1}$};

\draw[->, AccentTeal, semithick] (l0.south) -- (d1_0.north west);
\draw[->, AccentTeal, semithick] (l1.south) -- (d1_0.north east);
\draw[->, AccentTeal, semithick] (l2.south) -- (d1_1.north west);
\draw[->, AccentTeal, semithick] (l3.south) -- (d1_1.north east);
\draw[->, AccentTeal, semithick] (l4.south) -- (d1_2.north west);
\draw[->, AccentTeal, semithick] (l5.south) -- (d1_2.north east);
\draw[->, AccentTeal, semithick] (l6.south) -- (d1_3.north west);
\draw[->, AccentTeal, semithick] (l7.south) -- (d1_3.north east);

\node[anchor=east, font=\sffamily\scriptsize\bfseries, text=AccentTeal] at (0.2, -4.15) {Subtrees $\mathbf{H}_{\text{mid}}$};
\node[tree_node, minimum width=1.8cm] (d2_0) at (2.325, -4.15) {$\mathbf{h}^{(d^*)}_1$};
\node[tail_node, minimum width=1.8cm] (d2_1) at (8.525, -4.15) {$\mathbf{h}^{(d^*)}_{N_{\text{mid}}}$};

\draw[->, AccentTeal, semithick] (d1_0.south) -- (d2_0.north west);
\draw[->, AccentTeal, semithick] (d1_1.south) -- (d2_0.north east);
\draw[->, AccentTeal, semithick] (d1_2.south) -- (d2_1.north west);
\draw[->, AccentTeal, semithick] (d1_3.south) -- (d2_1.north east);

\draw[->, AccentPurple!80, thick, densely dashed, bend right=18] (l0.south west) to node[left, font=\sffamily\tiny\bfseries, text=AccentPurple] {4-Child Skip} (d2_0.west);
\draw[->, AccentPurple!80, thick, densely dashed, bend left=18] (l7.south east) to (d2_1.east);

\node[anchor=east, font=\sffamily\scriptsize\bfseries, text=AccentPurple] at (-0.3, -5.7) {Tree Root};
\node[root_node] (root) at (5.425, -5.5) {Root $\mathbf{h}_{\text{root}} \oplus \mathbf{h}_{\text{td}}$};
\draw[->, AccentTeal, thick] (d2_0.south) -- (root.north west);
\draw[->, AccentTeal, thick] (d2_1.south) -- (root.north east);

\draw[->, AccentPurple, thick, dotted, bend left=16] (root.west) to node[below left, font=\sffamily\tiny\bfseries, text=AccentPurple] {Top-Down Gate} (d2_0.south west);
\draw[->, AccentPurple, thick, dotted, bend right=16] (root.east) to node[below right, font=\sffamily\tiny\bfseries, text=AccentPurple] {Top-Down Gate} (d2_1.south east);

\draw[->, PrimaryBlue!85, thick, dashdotted] (-0.65, -1.35) -- (-1.65, -1.35) -- (-1.65, -5.5) -- (root.west);
\node[rotate=90, font=\sffamily\tiny\bfseries, text=PrimaryBlue] at (-1.9, -3.4) {Direct Leaf Avg/Max Pool \& Root Skip};

\node[mem_box, fill=SoftPurple, draw=AccentPurple, minimum width=2.4cm] (m_root) at (1.3, -6.9) {$\mathbf{r}_{\text{final}}$\\[-1pt]\tiny (1 Global Slot)};
\node[mem_box, fill=SoftTeal, draw=AccentTeal, minimum width=4.5cm] (m_mid) at (5.0, -6.9) {$\tilde{\mathbf{H}}_{\text{mid}} + \mathbf{E}_{\text{sub}}$\\[-1pt]\tiny ($\le 128$ Refined Subtree Slots)};
\node[mem_box, fill=SoftAmber, draw=AccentAmber, minimum width=4.2cm] (m_tail) at (9.6, -6.9) {$\mathbf{T}_{\text{tail}} + \mathbf{E}_{\text{tail}}$\\[-1pt]\tiny ($\le 64$ Fenwick Boundary Slots)};

\draw[->, AccentPurple, thick] (root.south) -- (m_root.north);
\draw[->, AccentTeal, thick] (d2_0.south) -- (m_mid.north);
\draw[->, AccentAmber, thick] (d1_3.south east) to[bend left=18] (m_tail.north east);
\draw[->, AccentAmber, thick] (d2_1.south east) -- (m_tail.north);

\node[anchor=east, font=\sffamily\scriptsize\bfseries, text=PrimaryBlue] at (-0.3, -6.9) {Memory $\mathbf{M}$};
\end{tikzpicture}
\caption{\textbf{Sequence-to-Graph-Tree Construction.} Adjacent context tokens are bound via shift-invariant bilinear interactions, causal Conv1D ($K=4$), and intra-chunk causal RoPE attention ($W=64$) to form leaf states $\mathbf{H}^{(0)}$. The hierarchy is recursively merged upward with 2-level quadruple shortcuts while extracting (i) rightmost Fenwick boundary tails $\mathbf{T}_{\text{tail}}$ at every depth, (ii) top-down refined subtree slots $\tilde{\mathbf{H}}_{\text{mid}}$ ($\le 128$ nodes) augmented with adaptive leaf pooling, and (iii) the global root $\mathbf{r}_{\text{final}}$.}
\label{fig:seq_to_tree}
\end{figure}

Given a raw context token sequence $\mathbf{c} = (c_1, c_2, \dots, c_L) \in \{0, \dots, V-1\}^L$ with hidden dimension $D = 1536$, direct pairwise binary pooling is vulnerable to phase-shift aliasing: an adjacent key-value pair $(c_{2k-1}, c_{2k})$ placed at an odd boundary $(c_{2k}, c_{2k+1})$ would be split across separate subtrees. To guarantee shift-invariant local binding before hierarchical graph construction, LimeTree executes a three-stage leaf preparation and multi-resolution graph-tree mapping pipeline, illustrated in Figure~\ref{fig:seq_to_tree}.

\paragraph{1. Shift-Invariant Bigram Leaf Binding and Causal Conv1D.}
First, token embeddings $\mathbf{H} = \text{Embed}(\mathbf{c}) \in \mathbb{R}^{L \times D}$ are normalized alongside their right-shifted neighbors $\mathbf{H}_{\text{next}} = [\mathbf{H}_{2:L}; \mathbf{0}]$ using FP32 root-mean-square normalization (\texttt{FP32RMSNorm}):
\begin{equation}
    \tilde{\mathbf{h}}_i = \text{RMSNorm}_{\text{curr}}(\mathbf{h}_i), \qquad \tilde{\mathbf{h}}_{i+1} = \text{RMSNorm}_{\text{next}}(\mathbf{h}_{i+1}).
\end{equation}
To bind every token to its immediate successor regardless of tree parity, a bilinear multiplicative interaction is computed and fused with a two-layer GELU binding network (\texttt{local\_bind}) and a causal depthwise separable 1D convolution (\texttt{CausalDepthwiseConv1d}, kernel size $K=4$):
\begin{align}
    \mathbf{b}_i &= \text{GELU}(\mathbf{W}_L \tilde{\mathbf{h}}_i) \odot (\mathbf{W}_R \tilde{\mathbf{h}}_{i+1}), \\
    \mathbf{u}_i &= \mathbf{W}_{\text{short}} \tilde{\mathbf{h}}_i + \text{MLP}_{\text{bind}}\big([\tilde{\mathbf{h}}_i \,\|\, \tilde{\mathbf{h}}_{i+1} \,\|\, \mathbf{b}_i]\big) + \tanh(\alpha_{\text{conv}}) \cdot \text{SiLU}\big(\text{Conv1D}_{K=4}(\tilde{\mathbf{H}})_i\big),
\end{align}
where $\mathbf{W}_{\text{short}}$ is initialized to the identity matrix $\mathbf{I}_D$ and $\alpha_{\text{conv}} \in \mathbb{R}$ is a learnable scalar gate initialized to $0.15$.

\paragraph{2. Linear-Time Intra-Chunk RoPE Leaf Enrichment.}
Before binary tree coarsening, the normalized bound states $\hat{\mathbf{U}} = \text{RMSNorm}_{\text{leaf}}(\mathbf{U})$ are partitioned into non-overlapping windows of width $W = 64$ (\texttt{LocalChunkAttentionBlock}). Within each chunk, causal multi-head self-attention ($H = 24$ heads, head dimension $d_h = 64$) with QK-RMSNorm, learnable log-scale temperature $\tau = \exp(\text{clamp}(s, -1.0, 3.2))$, rotary position embeddings (RoPE), and a SwiGLU feed-forward network enriches local entity representations in $\mathcal{O}(L \cdot W)$ time:
\begin{equation}
    \mathbf{H}^{(0)} = \text{RMSNorm}_{\text{post}}\Big(\hat{\mathbf{U}} + \tanh(\alpha_{\text{chunk}}) \cdot \text{LocalChunkAttn}_{W=64}(\hat{\mathbf{U}})\Big) \in \mathbb{R}^{L \times D}.
\end{equation}
The resulting tensor $\mathbf{H}^{(0)}$ serves as Level $d=0$ (the leaf layer) of the hierarchical graph-tree.

\paragraph{3. Multi-Scale Graph-Tree Topology and Memory Bank Assembly.}
Starting from $\mathbf{H}^{(0)}$ of length $L_0 = L$, the sequence is recursively coarsened over $D_{\text{tree}} = \lceil \log_2 L \rceil$ levels, where level $d$ contains $L_d = \lceil L_{d-1} / 2 \rceil$ nodes (padded with a zero node if $L_{d-1}$ is odd). Rather than retaining only the single root vector $\mathbf{H}^{(D_{\text{tree}})}$, \texttt{encode\_tree} extracts three complementary topological views from the graph-tree and concatenates them into a unified memory bank $\mathbf{M} \in \mathbb{R}^{M \times D}$ ($M \le 1 + 128 + 64 = 193$ slots):
\begin{enumerate}
    \item \textbf{Multi-Scale Fenwick Tail Bank ($\mathbf{T}_{\text{tail}} \in \mathbb{R}^{\le 64 \times D}$):} At each tree depth $d \in \{0, \dots, D_{\text{tree}}-1\}$, the rightmost 4 nodes $\mathbf{H}^{(d)}_{-4:}$ near the causal generation boundary are extracted and stamped with a depth embedding $\mathbf{e}^{\text{lvl}}_{\min(d, 15)}$ and branch-offset embedding $\mathbf{e}^{\text{br}}_{0:3}$:
    \begin{equation}
        \mathbf{T}^{(d)} = \mathbf{H}^{(d)}_{-4:} + \mathbf{e}^{\text{lvl}}_{\min(d, 15)} + \mathbf{e}^{\text{br}}_{0:3}.
    \end{equation}
    Concatenating $\{\mathbf{T}^{(d)}\}_{d=0}^{D_{\text{tree}}-1}$ (capped at 64 slots plus positional embeddings $\mathbf{E}_{\text{tail}}$) forms a logarithmic Fenwick-style horizon that provides high-resolution coverage of recent tokens and exponentially coarser summaries of distant tokens.
    \item \textbf{Mid-Level Subtree Slots with Direct Leaf Pooling ($\tilde{\mathbf{H}}_{\text{mid}} \in \mathbb{R}^{N_{\text{mid}} \times D}$):} The first tree level whose node count satisfies $L_{d^*} \le N_{\text{slots}} = 128$ is captured as the subtree memory tier $\mathbf{H}_{\text{mid}} = \mathbf{H}^{(d^*)}$. After top-down root conditioning ($\hat{\mathbf{H}}_{\text{mid}}$), it receives a direct residual skip connection from the leaf layer via dual adaptive average and max 1D pooling over $N_{\text{mid}}$ segments, followed by learned slot positional embeddings $\mathbf{E}_{\text{sub}}$:
    \begin{equation}
        \tilde{\mathbf{H}}_{\text{mid}} = \hat{\mathbf{H}}_{\text{mid}} + \tanh(\alpha_{\text{slot}}) \cdot \mathbf{W}_{\text{slot}}\,\text{RMSNorm}\!\left(\frac{1}{2}\text{AvgPool}_{N_{\text{mid}}}(\mathbf{H}^{(0)}) + \frac{1}{2}\text{MaxPool}_{N_{\text{mid}}}(\mathbf{H}^{(0)})\right) + \mathbf{E}_{\text{sub}}.
    \end{equation}
    \item \textbf{Express-Augmented Global Root ($\mathbf{r}_{\text{final}} \in \mathbb{R}^{1 \times D}$):} The top tree node $\mathbf{h}_{\text{root}}$ is augmented with a direct leaf-mean express projection $\mathbf{r}_{\text{leaf}} = \mathbf{W}_{\text{root}}\,\text{RMSNorm}(\frac{1}{L}\sum_{i=1}^L \mathbf{h}^{(0)}_i)$ and merged with the top-down summary vector $\mathbf{h}_{\text{td}}$ via $\mathbf{r}_{\text{final}} = \mathbf{h}_{\text{root}} + \mathbf{W}_{\text{merge}}[\mathbf{h}_{\text{root}} \,\|\, \mathbf{h}_{\text{td}}]$.
\end{enumerate}
Finally, the concatenated memory $\mathbf{M}_0 = [\mathbf{r}_{\text{final}}; \tilde{\mathbf{H}}_{\text{mid}}; \mathbf{T}_{\text{tail}}]$ undergoes all-to-all non-causal self-attention refinement (\texttt{GlobalMemoryRefineBlock}) to yield the final memory bank $\mathbf{M} = \text{RMSNorm}_{\text{mem}}(\text{GlobalRefine}(\mathbf{M}_0))$ and pooled context conditioning vector $\mathbf{c}_{\text{ctx}} = \text{MLP}_{\text{ctx}}(\mathbf{M}_{0,:} + \mathbf{r}_{\text{final}})$.

\subsection{GNN Tree Cell Architecture 1 (Wikitext)}

The \texttt{GraphTreeCell} module parameterizes message passing over the binary/quad-tree hierarchy through three directional operators: a convex-gated bottom-up binary combiner, a 2-hop quad-tree skip aggregator, and a root-conditioned top-down refinement pass.

\paragraph{Convex-Gated Bottom-Up Step.}
Given level states $\mathbf{H} \in \mathbb{R}^{L \times d}$ (padded to even length if $L$ is odd), left and right sibling nodes are partitioned into $\mathbf{H}_L = \mathbf{H}_{0::2}$ and $\mathbf{H}_R = \mathbf{H}_{1::2}$. Rather than relying on an unconstrained MLP, sibling aggregation is formulated as a learned convex combination modulated by relative node scores and cross-sibling interactions:
\begin{align}
    \mathbf{s}_L &= \text{MLP}_{\text{score}}(\text{LN}(\mathbf{H}_L)), \quad \mathbf{s}_R = \text{MLP}_{\text{score}}(\text{LN}(\mathbf{H}_R)), \\
    \mathbf{g} &= \sigma\left((\mathbf{s}_R - \mathbf{s}_L) + \tanh\left(\mathbf{W}_{\text{cross}}[\mathbf{H}_L \,\|\, \mathbf{H}_R]\right) + \mathbf{b}_{\text{override}}\right), \\
    \mathbf{H}_{\text{merged}} &= \mathbf{g} \odot \mathbf{H}_R + (\mathbf{1} - \mathbf{g}) \odot \mathbf{H}_L,
\end{align}
where $\sigma(\cdot)$ is the element-wise sigmoid and $\mathbf{b}_{\text{override}}$ is initialized to $0.15$. A gated post-merge feed-forward residual refines the combined parent representation:
\begin{equation}
    \mathbf{H}_{\text{parent}} = \mathbf{H}_{\text{merged}} + \tanh(\gamma_{\text{merge}}) \, \text{FFN}_{\text{merge}}(\mathbf{H}_{\text{merged}}).
\end{equation}

\paragraph{2-Hop Quad-Tree Shortcut.}
To mitigate gradient attenuation and information bottlenecks across binary tree depths, a 2-hop skip connection directly links groups of four grandchildren $\mathbf{H}_{\text{gc}} \in \mathbb{R}^{4 L_p \times d}$ to their corresponding grandparents $\mathbf{H}_{p} \in \mathbb{R}^{L_p \times d}$. Reshaping $\mathbf{H}_{\text{gc}}$ into $\mathbb{R}^{L_p \times 4 \times d}$, dual mean and max pooling capture both average context and salient features:
\begin{equation}
    \boldsymbol{\mu}_q = \frac{1}{4}\sum_{k=1}^4 \mathbf{H}_{\text{gc}}[:, k, :], \quad
    \boldsymbol{\nu}_q = \max_{k=1,\dots,4} \mathbf{H}_{\text{gc}}[:, k, :],
\end{equation}
which update the grandparent state via a tanh-gated linear projection:
\begin{equation}
    \mathbf{H}_{p} \leftarrow \mathbf{H}_{p} + \tanh(\gamma_{\text{quad}}) \, \mathbf{W}_{\text{quad}}\left[\text{LN}(\boldsymbol{\mu}_q) \,\|\, \text{LN}(\boldsymbol{\nu}_q)\right].
\end{equation}

\paragraph{Top-Down Node Refinement and Re-Aggregation.}
To propagate global sequence context back into intermediate subtree slots $\mathbf{H}_{\text{nodes}} \in \mathbb{R}^{M \times d}$, the root state $\mathbf{h}_{\text{root}}$ is broadcast and bound with each node through normalized bilinear multiplication:
\begin{align}
    \tilde{\mathbf{p}} &= \mathbf{W}_p \text{LN}(\mathbf{h}_{\text{root}}), \quad
    \tilde{\mathbf{n}}_j = \mathbf{W}_n \text{LN}(\mathbf{H}_{\text{nodes}, j}), \quad
    \mathbf{b}_j = \text{LN}(\tilde{\mathbf{n}}_j \odot \tilde{\mathbf{p}}), \\
    \mathbf{u}_j &= [\mathbf{H}_{\text{nodes}, j} \,\|\, \tilde{\mathbf{p}} \,\|\, \mathbf{b}_j] \in \mathbb{R}^{3d}, \\
    \hat{\mathbf{H}}_{\text{nodes}, j} &= \mathbf{H}_{\text{nodes}, j} + \sigma(\mathbf{W}_g \text{LN}(\mathbf{u}_j)) \odot \text{MLP}_{\text{td}}(\text{LN}(\mathbf{u}_j)).
\end{align}
The refined slots $\hat{\mathbf{H}}_{\text{nodes}}$ are subsequently reduced back to a single summary vector $\mathbf{h}_{\text{td}} \in \mathbb{R}^d$ via an lightweight bottom-up pass using dedicated score network $\text{MLP}_{\text{td\_score}}$ and cross-modulation matrix $\mathbf{W}_{\text{td\_cross}}$[cite: 1].

\subsection{Bidirectional Graph-Tree Neural Network Architecture (PG19)}
\label{subsec:gnn_arch}

\begin{figure}[htbp]
\centering
\begin{tikzpicture}[
    font=\sffamily\footnotesize,
    >=Stealth,
    box/.style={draw=SlateGray, fill=white, rounded corners=2.5pt, minimum height=0.58cm, inner sep=4pt, align=center},
    op/.style={draw=PrimaryBlue!85, fill=SoftBlue, rounded corners=3pt, minimum height=0.6cm, inner sep=4pt, align=center, font=\sffamily\scriptsize\bfseries},
    gate/.style={draw=AccentAmber!90, fill=SoftAmber, rounded corners=3pt, minimum height=0.6cm, inner sep=4pt, align=center, font=\sffamily\scriptsize\bfseries},
    norm/.style={draw=AccentTeal!90, fill=SoftTeal, rounded corners=2pt, minimum height=0.52cm, inner sep=3.5pt, align=center, font=\sffamily\scriptsize\bfseries},
    tdbox/.style={draw=AccentPurple!90, fill=SoftPurple, rounded corners=3pt, minimum height=0.6cm, inner sep=4pt, align=center, font=\sffamily\scriptsize\bfseries}
]

\begin{scope}[shift={(-2,0)}]
    \node[font=\sffamily\footnotesize\bfseries, text=PrimaryBlue] at (3.6, 4.9) {(a) Bottom-Up Convex Merge \& 4-Grandchild Shortcut};

    \node[box, minimum width=1.35cm, font=\sffamily\scriptsize] (gc0) at (0.4, 4.1) {$\mathbf{h}^{(d-1)}_{4k}$};
    \node[box, minimum width=1.35cm, font=\sffamily\scriptsize] (gc1) at (1.9, 4.1) {$\mathbf{h}^{(d-1)}_{4k+1}$};
    \node[box, minimum width=1.35cm, font=\sffamily\scriptsize] (gc2) at (3.4, 4.1) {$\mathbf{h}^{(d-1)}_{4k+2}$};
    \node[box, minimum width=1.35cm, font=\sffamily\scriptsize] (gc3) at (4.9, 4.1) {$\mathbf{h}^{(d-1)}_{4k+3}$};

    \node[op, minimum width=1.8cm] (hl) at (1.15, 2.95) {Left $\mathbf{h}^{(d)}_{2k}$};
    \node[op, minimum width=1.8cm] (hr) at (4.15, 2.95) {Right $\mathbf{h}^{(d)}_{2k+1}$};

    \draw[->, SlateGray, semithick] (gc0.south) -- (hl.north);
    \draw[->, SlateGray, semithick] (gc1.south) -- (hl.north);
    \draw[->, SlateGray, semithick] (gc2.south) -- (hr.north);
    \draw[->, SlateGray, semithick] (gc3.south) -- (hr.north);

    \node[box, font=\sffamily\scriptsize] (sl) at (-0.35, 1.8) {$\mathbf{s}_L = \text{Score}(\mathbf{h}_L)$};
    \node[box, font=\sffamily\scriptsize] (cm) at (2.9, 1.8) {$\mathbf{c}_{LR} = \tanh(\mathbf{W}_c[\mathbf{h}_L\|\mathbf{h}_R])$};
    \node[box, font=\sffamily\scriptsize] (sr) at (6.15, 1.8) {$\mathbf{s}_R = \text{Score}(\mathbf{h}_R)$};

    \draw[->, PrimaryBlue, semithick] (hl.south) -- (sl.north);
    \draw[->, PrimaryBlue, semithick] (hr.south) -- (sr.north);
    \draw[->, PrimaryBlue, semithick] (hl.south) -- (cm.north);
    \draw[->, PrimaryBlue, semithick] (hr.south) -- (cm.north);

    \node[gate, minimum width=4.6cm] (gbu) at (2.9, 0.75) {Convex Gate $\mathbf{g}_{\text{bu}} = \sigma\big((\mathbf{s}_R - \mathbf{s}_L) + \mathbf{c}_{LR} + \mathbf{b}_{\text{override}}\big)$};
    \draw[->, SlateGray, semithick] (sl.south) -- (gbu.north west);
    \draw[->, SlateGray, semithick] (cm.south) -- (gbu.north);
    \draw[->, SlateGray, semithick] (sr.south) -- (gbu.north east);

    \node[op, minimum width=4.6cm] (hmerge) at (2.9, -0.25) {$\mathbf{h}_{\text{merged}} = \mathbf{g}_{\text{bu}} \odot \mathbf{h}_R + (1 - \mathbf{g}_{\text{bu}}) \odot \mathbf{h}_L$};
    \draw[->, AccentAmber, thick] (gbu.south) -- (hmerge.north);

    \node[norm, minimum width=4.8cm] (hhat) at (2.9, -1.25) {$\hat{\mathbf{h}}^{(d+1)}_k = \text{RMSNorm}\big(\mathbf{h}_{\text{merged}} + \tanh(\beta_m)\text{FFN}(\mathbf{h}_{\text{merged}})\big)$};
    \draw[->, PrimaryBlue, thick] (hmerge.south) -- (hhat.north);

    \node[tdbox, minimum width=2.9cm] (quad) at (8.1, 0.75) {Quad Pool ($4\times$)\\[-1pt]\tiny $\text{Mean}(\mathbf{G}_k) \,\|\, \text{Max}(\mathbf{G}_k)$};
    \draw[->, AccentPurple, thick, densely dashed] (gc3.east) -| (quad.north);

    \node[norm, draw=PrimaryBlue, thick, minimum width=5.6cm] (hout) at (4.2, -2.35) {Parent Node $\mathbf{h}^{(d+1)}_k = \text{RMSNorm}\big(\hat{\mathbf{h}}^{(d+1)}_k + \tanh(\beta_q)\mathbf{W}_{\text{quad}}(\text{Quad})\big)$};
    \draw[->, AccentTeal, thick] (hhat.south) -- ([xshift=-1.3cm]hout.north);
    \draw[->, AccentPurple, thick, densely dashed] (quad.south) -- ([xshift=3.9cm]hout.north);

\end{scope}

\draw[BorderGray, thick, dashed] (7.8, -2.7) -- (7.8, 5.1);

\begin{scope}[shift={(8.8,0)}]
    \node[font=\sffamily\footnotesize\bfseries, text=AccentPurple] at (2.7, 4.9) {(b) Top-Down Root Refinement};

    \node[tdbox, minimum width=2.2cm] (r_in) at (1.2, 3.9) {Root $\mathbf{h}_{\text{root}}$};
    \node[op, minimum width=2.2cm] (m_in) at (4.1, 3.9) {Subtree $\mathbf{h}_m \in \mathbf{H}_{\text{mid}}$};

    \node[box, font=\sffamily\scriptsize, minimum width=2.1cm] (p_proj) at (1.2, 2.75) {$\mathbf{p} = \mathbf{W}_P\text{Norm}(\mathbf{h}_{\text{root}})$};
    \node[box, font=\sffamily\scriptsize, minimum width=2.1cm] (n_proj) at (4.1, 2.75) {$\mathbf{n}_m = \mathbf{W}_N\text{Norm}(\mathbf{h}_m)$};

    \draw[->, AccentPurple, semithick] (r_in.south) -- (p_proj.north);
    \draw[->, PrimaryBlue, semithick] (m_in.south) -- (n_proj.north);

    \node[tdbox, minimum width=4.6cm] (bi) at (2.65, 1.65) {Bilinear Fusion $\mathbf{z}_m = [\mathbf{h}_m \,\|\, \mathbf{p} \,\|\, \text{RMSNorm}(\mathbf{n}_m \odot \mathbf{p})]$};
    \draw[->, SlateGray, semithick] (p_proj.south) -- (bi.north);
    \draw[->, SlateGray, semithick] (n_proj.south) -- (bi.north);

    \node[gate, minimum width=2.2cm] (td_g) at (1.3, 0.45) {Gate $\mathbf{g}^{\text{td}}_m = \sigma(\mathbf{z}_m)$};
    \node[box, font=\sffamily\scriptsize, minimum width=2.3cm] (td_u) at (4.0, 0.45) {Update $\text{MLP}_{\text{td}}(\mathbf{z}_m)$};

    \draw[->, AccentPurple, semithick] (bi.south west) -- (td_g.north);
    \draw[->, AccentPurple, semithick] (bi.south east) -- (td_u.north);

    \node[norm, minimum width=4.8cm] (m_ref) at (2.65, -0.8) {Refined Slot $\hat{\mathbf{h}}_m = \text{RMSNorm}\big(\mathbf{h}_m + \mathbf{g}^{\text{td}}_m \odot \text{MLP}_{\text{td}}(\mathbf{z}_m)\big)$};
    \draw[->, AccentAmber, semithick] (td_g.south) -- (m_ref.north);
    \draw[->, SlateGray, semithick] (td_u.south) -- (m_ref.north);

    \node[tdbox, minimum width=4.8cm] (td_sum) at (2.65, -2.15) {Secondary Convex Tree Reduction $\longrightarrow \mathbf{h}_{\text{td}} \in \mathbb{R}^D$};
    \draw[->, AccentTeal, thick] (m_ref.south) -- (td_sum.north);
\end{scope}

\end{tikzpicture}
\caption{\textbf{Architecture of the GNN.} \textbf{(a)} During the bottom-up pass, sibling pairs $(\mathbf{h}_L, \mathbf{h}_R)$ are combined via a learned differential convex gate $\mathbf{g}_{\text{bu}}$ with a right-branch recency prior $\mathbf{b}_{\text{override}}$, followed by a $\tanh$-gated FFN and multi-hop 4-grandchild mean/max skip connections every two tree levels. \textbf{(b)} During the top-down pass, the global root $\mathbf{h}_{\text{root}}$ modulates mid-level subtree slots $\mathbf{H}_{\text{mid}}$ through a multiplicative bilinear gate before a secondary convex tree sweep produces the global summary $\mathbf{h}_{\text{td}}$.}
\label{fig:gnn_cell}
\end{figure}
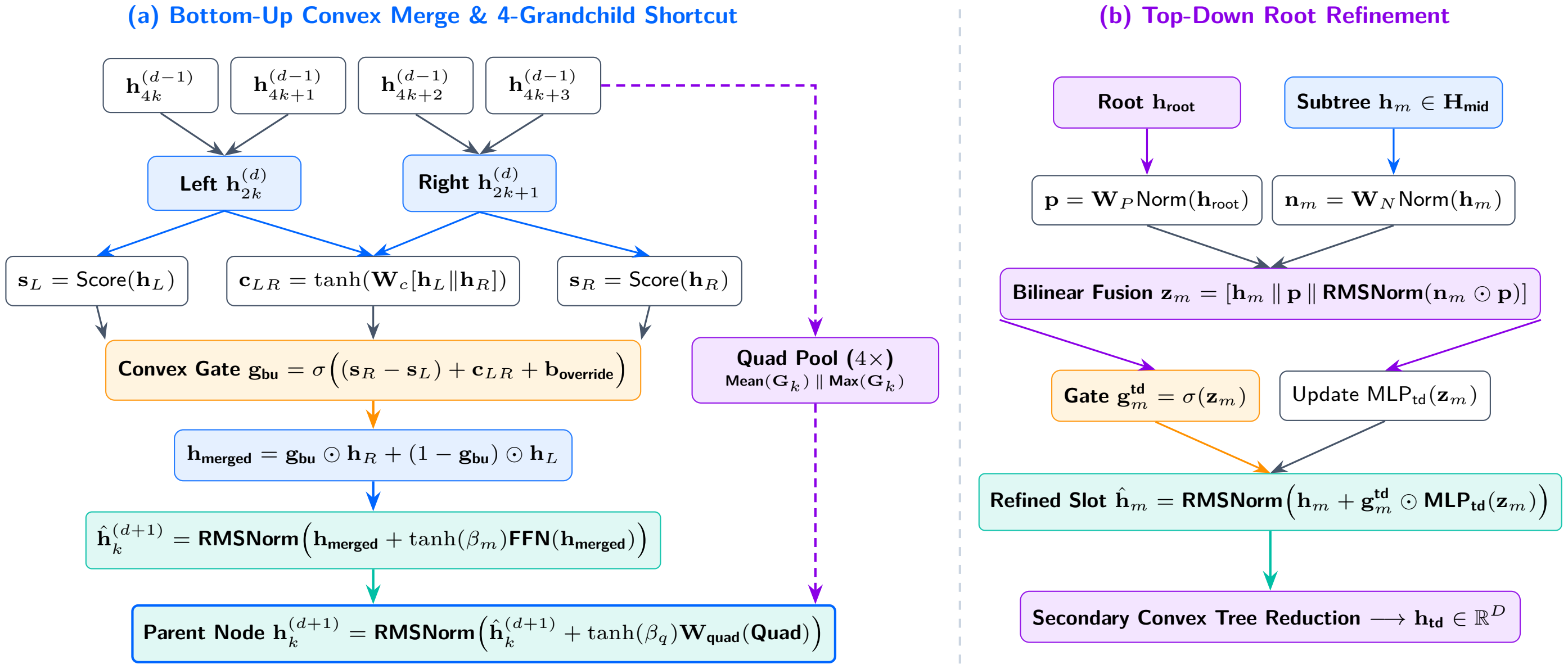

Standard tree-structured networks suffer from gradient vanishing or norm explosion over logarithmic depths ($\sim 12$ levels at $L=4096$) and struggle when downstream queries require top-down global context to disambiguate local subtree representations. To overcome both limitations, \texttt{GraphTreeCell} implements a stabilized, three-phase bidirectional message-passing GNN across the hierarchical graph-tree, depicted in Figure~\ref{fig:gnn_cell}.

\paragraph{Phase 1: Convex-Gated Bottom-Up Message Passing (\texttt{bottom\_up\_step}).}
At any tree level $d$, sibling node pairs $(\mathbf{h}_L, \mathbf{h}_R) = (\mathbf{H}^{(d)}_{2k}, \mathbf{H}^{(d)}_{2k+1}) \in \mathbb{R}^D$ are merged into a parent node at level $d+1$. Rather than using an unconstrained linear projection that can exponentially amplify activation norms across depth, \texttt{GraphTreeCell} computes a competitive differential score between siblings alongside a cross-modulation term:
\begin{align}
    \mathbf{s}_L &= \text{MLP}_{\text{score}}(\text{RMSNorm}(\mathbf{h}_L)), \qquad
    \mathbf{s}_R = \text{MLP}_{\text{score}}(\text{RMSNorm}(\mathbf{h}_R)), \\
    \mathbf{c}_{LR} &= \tanh\!\big(\mathbf{W}_{\text{cross}} [\mathbf{h}_L \,\|\, \mathbf{h}_R] + \mathbf{b}_{\text{cross}}\big), \\
    \mathbf{g}_{\text{bu}} &= \sigma\!\big((\mathbf{s}_R - \mathbf{s}_L) + \mathbf{c}_{LR} + \mathbf{b}_{\text{override}}\big) \in (0, 1)^D,
\end{align}
where $\text{MLP}_{\text{score}}$ is a two-layer GELU network and $\mathbf{b}_{\text{override}} \in \mathbb{R}^D$ is a learnable recency prior initialized to $+0.15$ so that newer (right-branch) updates gently dominate when sibling salience is tied. The parent representation is formed via an element-wise convex combination followed by a $\tanh$-gated residual FFN and level-wise RMS stabilization:
\begin{align}
    \mathbf{h}_{\text{merged}} &= \mathbf{g}_{\text{bu}} \odot \mathbf{h}_R + (1 - \mathbf{g}_{\text{bu}}) \odot \mathbf{h}_L, \\
    \hat{\mathbf{h}}^{(d+1)}_k &= \text{RMSNorm}_{\text{lvl}}\!\Big(\mathbf{h}_{\text{merged}} + \tanh(\beta_{\text{merge}}) \cdot \text{MLP}_{\text{merge}}(\mathbf{h}_{\text{merged}})\Big).
\end{align}
Because $\mathbf{h}_{\text{merged}}$ lies strictly in the channel-wise convex hull of $(\mathbf{h}_L, \mathbf{h}_R)$ and $\tanh(\beta_{\text{merge}})$ bounds the non-linear correction (initialized at $\beta_{\text{merge}} = 0.15$), forward activations and backward gradients remain norm-preserving across arbitrary tree depths.

\paragraph{Phase 2: Multi-Hop Quadruple Grandchild Shortcuts (\texttt{quad\_shortcut\_step}).}
Every two tree levels ($d-1 \to d+1$), each parent node $\hat{\mathbf{h}}^{(d+1)}_k$ spans a 4-tuple of grandchildren $\mathbf{G}_k = \{\mathbf{h}^{(d-1)}_{4k+j}\}_{j=0}^3 \in \mathbb{R}^{4 \times D}$. To create high-bandwidth gradient highways that halve the effective path length of the binary tree, \texttt{GraphTreeCell} aggregates both the mean and channel-wise maximum across the four grandchildren:
\begin{align}
    \boldsymbol{\mu}_k &= \frac{1}{4}\sum_{j=0}^{3} \mathbf{h}^{(d-1)}_{4k+j}, \qquad
    \boldsymbol{\nu}_k = \max_{j \in \{0,1,2,3\}} \mathbf{h}^{(d-1)}_{4k+j}, \\
    \mathbf{h}^{(d+1)}_k &= \text{RMSNorm}_{\text{lvl}}\!\Big(\hat{\mathbf{h}}^{(d+1)}_k + \tanh(\beta_{\text{quad}}) \cdot \mathbf{W}_{\text{quad}} \big[\text{RMSNorm}_{\text{mean}}(\boldsymbol{\mu}_k) \,\|\, \text{RMSNorm}_{\text{max}}(\boldsymbol{\nu}_k)\big]\Big),
\end{align}
where $\boldsymbol{\nu}_k$ preserves sharp, sparse key-value spikes while $\boldsymbol{\mu}_k$ preserves distributional semantics.

\paragraph{Phase 3: Top-Down Bilinear Root Conditioning and Re-Aggregation (\texttt{top\_down\_refine\_nodes}).}
Once upward message passing reaches $\mathbf{h}_{\text{root}}$, information flows top-down to contextualize the mid-level subtree nodes $\mathbf{H}_{\text{mid}} = \{\mathbf{h}_m\}_{m=1}^{N_{\text{mid}}}$. The normalized root and subtree nodes are projected into a shared bilinear interaction space:
\begin{align}
    \mathbf{p} &= \mathbf{W}^{\text{td}}_P\,\text{RMSNorm}_P(\mathbf{h}_{\text{root}}), \qquad
    \mathbf{n}_m = \mathbf{W}^{\text{td}}_N\,\text{RMSNorm}_N(\mathbf{h}_m), \\
    \mathbf{z}_m &= \big[\mathbf{h}_m \,\|\, \mathbf{p} \,\|\, \text{RMSNorm}_{\text{bi}}(\mathbf{n}_m \odot \mathbf{p})\big] \in \mathbb{R}^{3D}.
\end{align}
A sigmoid gate $\mathbf{g}^{\text{td}}_m = \sigma(\mathbf{W}_G\,\text{RMSNorm}(\mathbf{z}_m))$ selectively injects the top-down update:
\begin{equation}
    \hat{\mathbf{h}}_m = \text{RMSNorm}_{\text{lvl}}\!\Big(\mathbf{h}_m + \mathbf{g}^{\text{td}}_m \odot \text{MLP}_{\text{td}}(\mathbf{z}_m)\Big).
\end{equation}
Finally, the refined mid-level nodes $\{\hat{\mathbf{h}}_m\}_{m=1}^{N_{\text{mid}}}$ are reduced back up to a single vector $\mathbf{h}_{\text{td}} \in \mathbb{R}^D$ via a secondary parameter-specialized convex tree reduction ($\text{MLP}_{\text{td\_score}}, \mathbf{W}_{\text{td\_cross}}$), closing the bidirectional loop before the causal Transformer decoder cross-attends to $\mathbf{M}$.

\begin{table}[htbp]
\centering
\small
\caption{Summary of LimeTree Hierarchical Graph-Tree Hyperparameters ($D = 1536$).}
\label{tab:hyperparams}
\begin{tabular}{@{}llp{6.8cm}@{}}
\toprule
\textbf{Component / Module} & \textbf{Configuration} & \textbf{Role in Graph-Tree \& GNN Pipeline} \\
\midrule
Leaf Bigram \& Conv1D & $K=4$, $\alpha_{\text{conv}}=0.15$ & Shift-invariant odd/even token pair binding \\
\texttt{LocalChunkAttentionBlock} & $W=64$, $H=24$, $d_h=64$ & $\mathcal{O}(L \cdot W)$ intra-chunk RoPE leaf enrichment \\
\texttt{GraphTreeCell} (Bottom-Up) & $\mathbf{b}_{\text{override}}=0.15$, $\beta_m=0.15$ & Norm-preserving convex sibling interpolation \\
Quadruple Skip Connection & $4\to 1$ Mean$\|$Max, $\beta_q=0.15$ & Halves tree path length; preserves sparse KV spikes \\
Top-Down Root Refinement & $3D \to D$ bilinear gate & Broadcasts global root context into $\mathbf{H}_{\text{mid}}$ \\
Memory Bank $\mathbf{M}$ & $1 + 128 + 64 = 193$ slots & Global root $\mathbf{r}_{\text{final}}$, subtrees $\tilde{\mathbf{H}}_{\text{mid}}$, Fenwick tails $\mathbf{T}_{\text{tail}}$ \\
\bottomrule
\end{tabular}
\end{table}

\subsection{Multi-GPU Distributed Training}
\label{sec:multi_gpu_ddp}

To scale long-context training across $R=4$ NVIDIA Quadro RTX 8000 GPUs ($4 \times 48\,\text{GB}$ GDDR6, Turing SM~7.5), the optimization pipeline employs Distributed Data Parallelism (DDP) coupled with native FP16 Tensor Core execution and dynamic loss scaling. Rather than streaming batches over PCIe during training, the pre-tokenized PG-19 corpus ($2 \times 10^9$ BPE tokens) is shifted by $-32768$ and stored losslessly as signed 16-bit integers (\texttt{int16}, $4\,\text{GB}$) directly within each GPU's local VRAM. Each rank $r \in \{0, \dots, R-1\}$ initializes an independent on-device pseudo-random number generator with seed
\begin{equation}
    s_r = s_0 + 10007\,r,
\end{equation}
ensuring mutually disjoint document spans and synthetic associative recall (KV) sequences across all devices with zero host-to-device transfer overhead.

Because Turing SM~7.5 architectures provide $130.5\,\text{TFLOPS}$ of FP16 Tensor Core throughput per device without native \texttt{bfloat16} hardware, forward and backward passes execute under mixed-precision FP16 while maintaining FP32 master weights, FP32 RMSNorm statistics, and FP32 chunked cross-entropy logits. To prevent FP16 logit drift and gradient underflow, the language modeling objective incorporates an auxiliary log-partition penalty ($z$-loss) with coefficient $\lambda_z = 10^{-4}$ alongside dynamic loss scaling factor $\sigma_t$:
\begin{equation}
    \mathcal{L}_{\text{LM}} = \frac{w_{\text{gen}}\,\mathcal{L}_{\text{CE}}^{\text{gen}} + w_{\text{tail}}\,\mathcal{L}_{\text{CE}}^{\text{tail}}}{w_{\text{gen}} + w_{\text{tail}}} + \lambda_z \frac{1}{T}\sum_{i=1}^{T} \left(\log \sum_{v=1}^{V} e^{z_{i,v}}\right)^2.
\end{equation}

\begin{figure*}[htbp]
\centering
\begin{tikzpicture}[
    font=\sffamily\scriptsize,
    >=Latex,
    node distance=4mm and 6mm,
    gpu_box/.style={
        draw=black!75, rounded corners=3pt, fill=gray!6,
        minimum width=3.6cm, minimum height=5.5cm, align=center
    },
    vram_block/.style={
        draw=blue!60!black, fill=blue!10, rounded corners=2pt,
        minimum width=3.2cm, text width=3.2cm, minimum height=0.85cm,
        inner sep=3pt, align=center
    },
    nosync_block/.style={
        draw=orange!75!black, fill=orange!12, rounded corners=2pt,
        minimum width=3cm, minimum height=1.15cm, align=center
    },
    sync_block/.style={
        draw=teal!80!black, fill=teal!14, rounded corners=2pt,
        minimum width=3cm, minimum height=0.95cm, align=center
    },
    nccl_bus/.style={
        draw=purple!80!black, fill=purple!12, rounded corners=3pt,
        minimum width=15.6cm, minimum height=0.75cm, align=center, font=\sffamily\scriptsize\bfseries
    },
    step_block/.style={
        draw=black!80, fill=green!12, rounded corners=2pt,
        minimum width=15.6cm, minimum height=0.7cm, align=center, font=\sffamily\scriptsize\bfseries
    },
    arrow/.style={->, thick, draw=black!75},
    sync_arrow/.style={<->, very thick, draw=purple!80!black}
]

\foreach \r/\x in {0/0, 1/4.0, 2/8.0, 3/12.0} {
    \node[gpu_box] (gpu\r) at (\x, 0) {};
    \node[anchor=north, font=\sffamily\scriptsize\bfseries] at (gpu\r.north)
        {GPU \r: Quadro RTX 8000};

    \node[vram_block] (vram\r) at (\x, 1.55)
        {Resident PG-19 (\texttt{int16}, 4\,GB)\\Disjoint RNG ($s_{\r} = s_0 + 10007\cdot\r$)};

    \node[nosync_block] (nosync\r) at (\x, -0.1)
        {\textbf{Local Accum. (\texttt{no\_sync})}\\
         KV Micro-Batches ($1\dots K$)\\
         LM Micro-Batches ($1\dots M-1$)};

    \node[sync_block] (sync\r) at (\x, -1.55)
        {\textbf{Final LM Micro-Batch ($M$)}\\
         FP16 Fwd/Bwd + Tree Encoder};

    \draw[arrow] (vram\r) -- (nosync\r);
    \draw[arrow] (nosync\r) -- (sync\r);
}

\node[nccl_bus] (nccl) at (6.0, -3.35)
    {Overlapped NCCL Ring-AllReduce (64\,MB Gradient Buckets, FP16 $\rightarrow$ FP32 Unscale)};

\foreach \r in {0, 1, 2, 3} {
    \draw[sync_arrow] (sync\r.south) -- (sync\r.south |- nccl.north);
}

\node[step_block] (opt) at (6.0, -4.6)
    {Lockstep Global $\ell_2$ Grad Clip ($1.0$ / $0.75$ Tree) $\rightarrow$ Fused AdamW Step $\rightarrow$ In-Place EMA Update ($\beta=0.9995$)};

\draw[arrow, very thick] (nccl.south) -- (opt.north);

\end{tikzpicture}
\caption{Distributed training dataflow across four NVIDIA Quadro RTX 8000 GPUs. Each device samples disjoint sequences from a local 4\,GB \texttt{int16} VRAM cache and accumulates gradients locally under \texttt{no\_sync()} before overlapping bucketed AllReduce communication with the final micro-batch backward pass.}
\label{fig:rtx8000_ddp_pipeline}
\end{figure*}
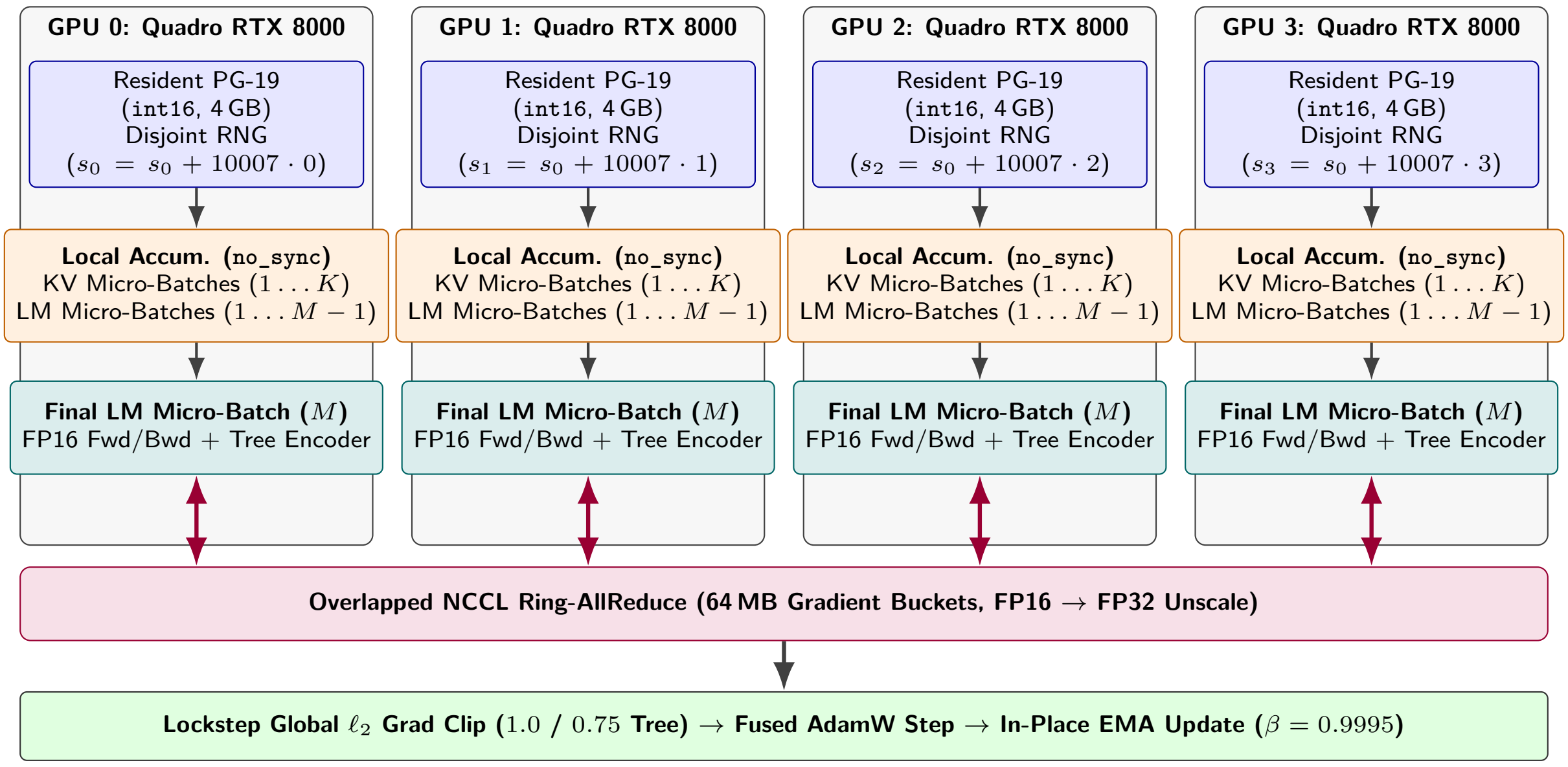

To minimize inter-GPU communication overhead during dual-task optimization, gradient synchronization is decoupled from micro-batch accumulation (Algorithm~\ref{alg:ddp_step} and Figure~\ref{fig:rtx8000_ddp_pipeline}). At each step $t$, all $K$ associative recall micro-batches and the first $M-1$ PG-19 language modeling micro-batches execute within a local accumulation context (\texttt{no\_sync}), accumulating scaled gradients $\sigma_t g_r$ in local VRAM without invoking collective communications. Synchronization is enabled exclusively on the final language modeling micro-batch $M$, overlapping backward-pass gradient computation with $64\,\text{MB}$ bucketed NCCL Ring-AllReduce across the four GPUs:
\begin{equation}
    \bar{g} = \frac{1}{\sigma_t R} \sum_{r=0}^{R-1} \left( \sum_{k=1}^{K} \frac{\lambda_{\text{KV}}}{K} \sigma_t \nabla_\theta \mathcal{L}_{\text{KV}}^{(r,k)} + \sum_{m=1}^{M} \frac{1}{M} \sigma_t \nabla_\theta \mathcal{L}_{\text{LM}}^{(r,m)} \right).
\end{equation}
Following gradient unscaling and separate $\ell_2$-norm clipping for the hierarchical tree cell ($\tau_{\text{tree}} = 0.75$) and global parameters ($\tau_{\text{global}} = 1.0$), all ranks execute a lockstep fused AdamW update and an in-place Exponential Moving Average (EMA) update $\theta_{\text{EMA}} \leftarrow \beta_t \theta_{\text{EMA}} + (1 - \beta_t)\theta$.

\begin{algorithm}[t]
\caption{4-GPU DDP Dual-Task Optimization Step with Overlapped AllReduce}
\label{alg:ddp_step}
\begin{algorithmic}[1]
\Require Model parameters $\theta$, EMA weights $\theta_{\text{EMA}}$, loss scale $\sigma_t$, rank $r \in \{0,1,2,3\}$
\State $\nabla_\theta \leftarrow 0$ \Comment{Reset local gradient buffers}
\State \textbf{with} \texttt{ddp\_model.no\_sync()} \textbf{do} \Comment{Phase 1: Local KV accumulation}
\For{$k = 1$ \textbf{to} $K$}
    \State Sample KV batch $\mathcal{B}_{\text{KV}}^{(r,k)}$ using rank generator $\mathcal{G}(s_r)$
    \State Compute $\mathcal{L}_{\text{KV}}^{(r,k)}$ in FP16 autocast with FP32 logits
    \State Accumulate local gradients: $\nabla_\theta \mathrel{+}= \sigma_t \nabla_\theta \left(\frac{\lambda_{\text{KV}}}{K}\mathcal{L}_{\text{KV}}^{(r,k)}\right)$
\EndFor
\State \textbf{end with}
\For{$m = 1$ \textbf{to} $M$} \Comment{Phase 2: PG-19 LM accumulation \& sync}
    \State Sample slice $\mathcal{B}_{\text{LM}}^{(r,m)}$ from VRAM-resident \texttt{int16} corpus via $\mathcal{G}(s_r)$
    \If{$m < M$}
        \State Compute $\mathcal{L}_{\text{LM}}^{(r,m)}$ and accumulate $\nabla_\theta \mathrel{+}= \sigma_t \nabla_\theta \left(\frac{1}{M}\mathcal{L}_{\text{LM}}^{(r,m)}\right)$ under \texttt{no\_sync()}
    \Else
        \State Compute $\mathcal{L}_{\text{LM}}^{(r,M)}$ and trigger overlapped 64\,MB bucketed Ring-AllReduce:
        \State $\nabla_\theta \leftarrow \frac{1}{R}\sum_{r=0}^{R-1} \left(\nabla_\theta + \sigma_t \nabla_\theta \left(\frac{1}{M}\mathcal{L}_{\text{LM}}^{(r,M)}\right)\right)$
    \EndIf
\EndFor
\State Unscale gradients: $\bar{g} \leftarrow \nabla_\theta / \sigma_t$ \Comment{Phase 3: Lockstep optimizer \& EMA step}
\State Clip norms: $\text{ClipNorm}(\bar{g}_{\text{tree}}, 0.75)$, $\text{ClipNorm}(\bar{g}, 1.0)$
\If{$\text{isfinite}(\|\bar{g}\|_2)$}
    \State $\theta \leftarrow \text{FusedAdamW}(\theta, \bar{g}, \eta_t)$
    \State $\theta_{\text{EMA}} \leftarrow \beta_t \theta_{\text{EMA}} + (1 - \beta_t)\theta$
\EndIf
\State Update dynamic loss scaler $\sigma_{t+1} \leftarrow \text{GradScaler}(\sigma_t, \|\bar{g}\|_2)$
\end{algorithmic}
\end{algorithm}

\subsection{Training Setup}
\label{sec:training_setup}

We evaluate our architecture across two complementary benchmarks designed to test medium-range language modeling and long-range multi-scale context extrapolation: \textbf{WikiText-103} and \textbf{PG-19}. In both regimes, the model is trained jointly on autoregressive language modeling (LM) and a synthetic associative Key-Value (KV) retrieval task to encourage robust long-horizon memory routing alongside local token prediction.

\paragraph{WikiText-103 Configuration.}
For the WikiText-103 benchmark, the language modeling objective is trained at a fixed context length of $L_{\text{LM}} = 512$ tokens. To supervise associative recall across varying memory horizons up to the full context window, the co-trained KV retrieval task samples sequences across four dyadic context lengths:
\[
    L_{\text{KV}}^{\text{WT103}} \in \{64, 128, 256, 512\}.
\]
Training is conducted in a data-parallel configuration across $4\times$ NVIDIA Quadro RTX 8000 GPUs ($48\text{ GB}$ GDDR6 VRAM per device; $192\text{ GB}$ aggregate cluster memory).

\paragraph{PG-19 Multi-Scale Length Curriculum.}
To scale the model to book-length narratives on PG-19 without incurring the prohibitive cost of training exclusively at the maximum sequence length throughout the entire optimization trajectory, we employ a stochastic multi-scale sequence-length curriculum. At each language modeling step (or epochal allocation), the training sequence length is drawn from the weighted multiset:
\[
    L_{LM}^{PG19} = [1024, 2048, 2048, 4096, 4096, 4096].
\]
This sampling schedule induces a categorical distribution over context windows with probabilities:
\[
    P(L_{\text{LM}} = 1024) = \frac{1}{6} \;(16.7\%), \quad
    P(L_{\text{LM}} = 2048) = \frac{2}{6} \;(33.3\%), \quad
    P(L_{\text{LM}} = 4096) = \frac{3}{6} \;(50.0\%),
\]
yielding an expected training context length of $\mathbb{E}[L_{\text{LM}}] = 2{,}901.3$ tokens. Half of all language modeling updates occur at the full $4{,}096$-token horizon, while the remaining updates at $1{,}024$ and $2{,}048$ tokens regularize intermediate positional representations and accelerate gradient throughput. Concurrently, the auxiliary KV retrieval task is expanded across seven dyadic scales spanning two orders of magnitude:
\[
    L_{KV}^{PG19} = \{64, 128, 256, 512, 1024, 2048, 4096\}.
\]
The PG-19 model is trained on a single Nvidia Quadro RTX 6000 Pro Blackwell Server Edition (BSE).

\paragraph{Training Setup and Optimization.}
Language modeling and KV retrieval tasks are jointly trained by alternating batches across sampled sequence lengths while adjusting gradient accumulation to maintain a constant global token budget per step. Models are trained in mixed precision using AdamW with a cosine learning rate schedule, linear warmup, weight decay, and gradient clipping. Table~\ref{tab:training_setup} summarizes the configurations for both settings.

\begin{table}[ht]
\centering
\caption{Training and evaluation configurations for the WikiText-103 and PG-19 models.}
\label{tab:training_setup}
\vspace{0.4em}
\small
\begin{tabularx}{\textwidth}{@{}l X X@{}}
\toprule
\textbf{Configuration} & \textbf{WikiText-103 Model} & \textbf{PG-19 Long-Context Model} \\
\midrule
LM Training Context ($L_{\text{LM}}$) & Fixed: $512$ & Multi-scale: $\{1024, 2048, 4096\}$ \\
LM Sampling Weights & $1.0$ & $1 : 2 : 3$ \\
KV Training Lengths ($L_{\text{KV}}$) & $\{64, 128, 256, 512\}$ & $\{64, 128, \dots, 4096\}$ \\
Evaluation Lengths ($L_{\text{eval}}$) & $\{64, 128, \dots, 65536\}$ ($128\times$ extrap.) & $\{64, 128, \dots, 524288\}$ ($128\times$ extrap.) \\
Hardware (DDP) & $4\times$ Quadro RTX 8000 & $1\times$ Quadro RTX 6000 Pro BSE \\
\bottomrule
\end{tabularx}
\end{table}

\section{Results and Discussion}

\subsection{Wikitext 103 results}

\begin{figure}[htbp]
    \centering
    \includegraphics[width=1\linewidth]{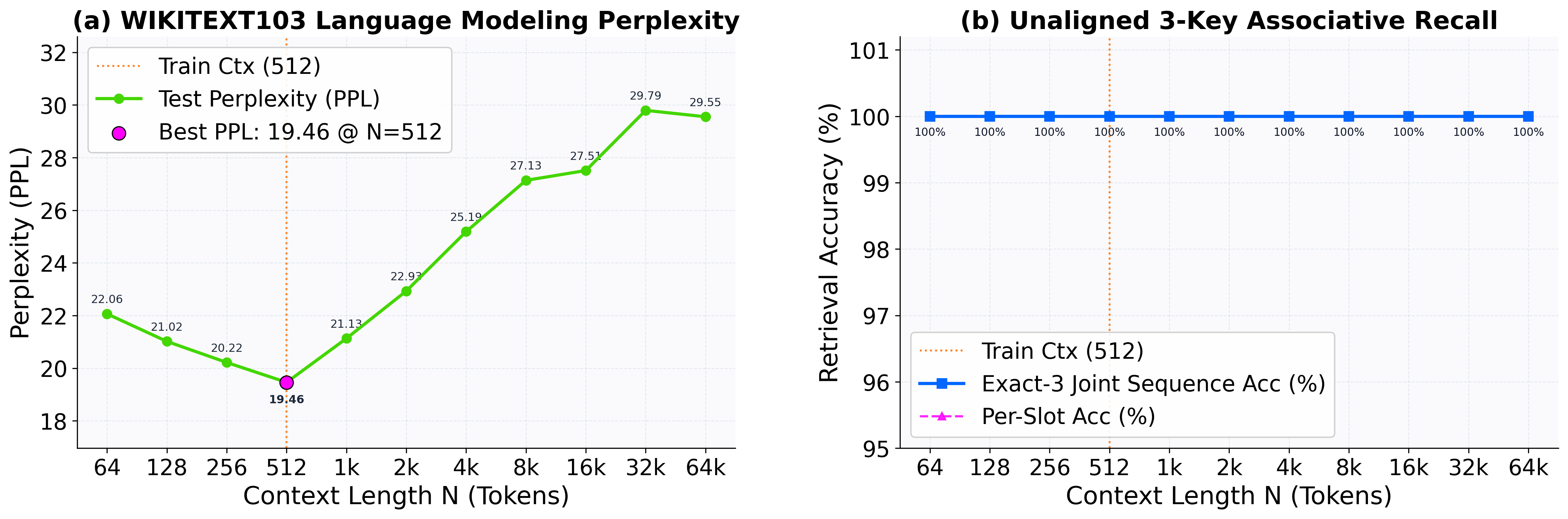}
    \caption{Performance of LimeTree 381 on Wikitext 103. (a) shows the excellent extrapolation capabilities, where the best perplexity score is achieved at the training context length 512. (b) shows the performance on the KV retrieval task.}
    \label{fig:wikitext_main_results}
\end{figure}

To evaluate language modeling performance, a $381\text{M}$-parameter model ($316\text{M}$ non-embedding) was trained on the $103\text{M}$-token WikiText-103 dataset with a maximum context length of $512$. Evaluated at this training horizon, the model achieves a test perplexity (PPL) of $19.46$. More notably, it generalizes zero-shot well beyond its $512$-token training window (Figure~\ref{fig:wikitext_main_results}(a)). Unlike standard Transformers, which typically break down once pushed past their training context (as reported in [\cite{deletang2023neural, press2022train}]; not tested here), the recursive tree encoder extends to $65{,}536$ tokens—a $128\times$ increase—without collapsing. Perplexity rises gradually across log-spaced horizons, moving from $25.19$ at $4{,}096$ tokens to $29.55$ at $65{,}536$. With the exact same checkpoint weights, the model also carries this extrapolation over to key-value retrieval, maintaining $100\%$ exact-match accuracy from $64$ up to $65{,}536$ tokens (Figure~\ref{fig:wikitext_main_results}(b)). Even when compressing sequences orders of magnitude longer than those seen during training, the hierarchical memory slots continue to preserve specific associations.

\begin{figure}
    \centering
    \includegraphics[width=1\linewidth]{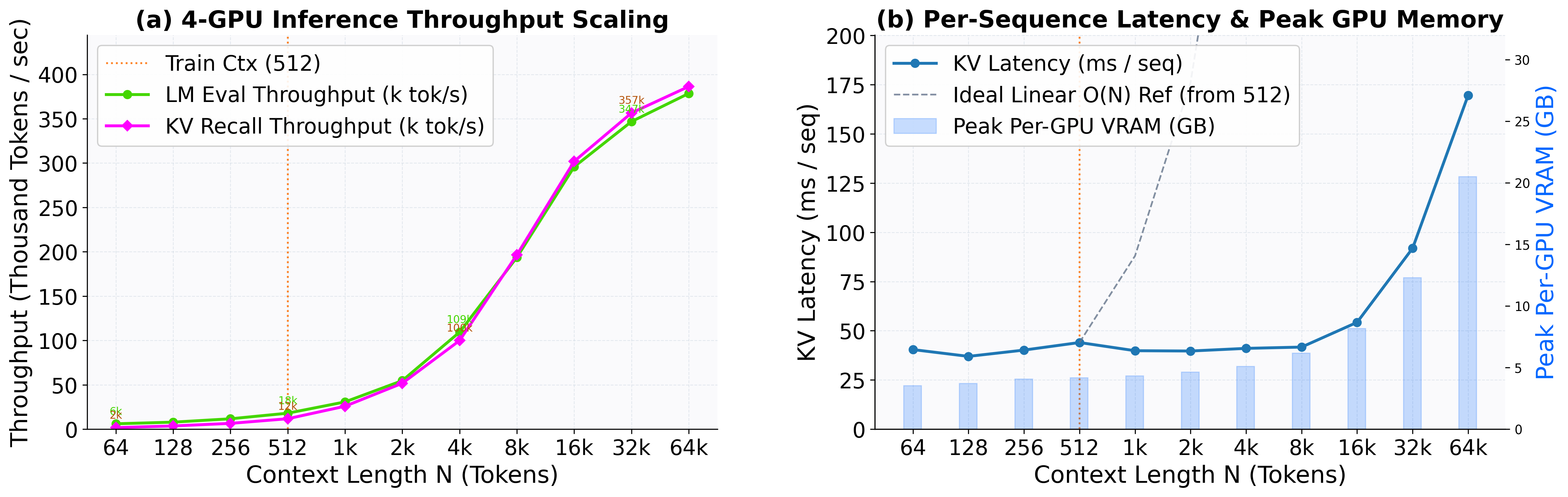}
    \caption{Processing speed of LimeTree 381 on Wikitext 103. (a) shows the tokens per second in both the language modeling (LM) and key-value retrieval (KV) task. (b) shows the latency in the key-value retrieval task as well as the peak GPU VRAM usage in the language modeling task. The orange dashed line indicates the training context length of $512$.}
    \label{fig:wikitext_tokens_per_second}
\end{figure}

Because LimeTree compresses long histories through a hierarchical tree rather than full pairwise attention, both its compute and memory demands scale linearly with sequence length. In practice, this fundamentally changes how inference behaves as prompts get longer. As shown in Figure~\ref{fig:wikitext_tokens_per_second}(a), whereas standard Transformers hit a quadratic bottleneck that causes processing speed to fall off at long horizons, LimeTree's throughput climbs steadily across log-spaced context lengths—rising from roughly $65\text{k}\,\text{tokens/s}$ at $512$ tokens and $100\text{k}\,\text{tokens/s}$ at $1{,}024$, up to nearly $400\text{k}\,\text{tokens/s}$ at $65{,}536$ tokens. At shorter lengths, runtime is bounded by the fixed-size causal decoder tail and kernel overhead; as the input grows, the lightweight $\mathcal{O}(N)$ tree encoder amortizes that constant cost and keeps the GPU saturated, ingesting a full $32\text{k}$-token context in just over $100\,\text{ms}$. Figure~\ref{fig:wikitext_tokens_per_second}(b) shows the corresponding memory behavior: peak VRAM consumption grows linearly with context length rather than quadratically. Because older context is folded into fixed-capacity tree slots while the causal decoder operates over a constant $256$-token window, the memory footprint increases at a predictable, bounded rate per token out to $32{,}768$ tokens, avoiding the memory-capacity walls that limit standard self-attention.

\begin{figure}
    \centering
    \includegraphics[width=1\linewidth]{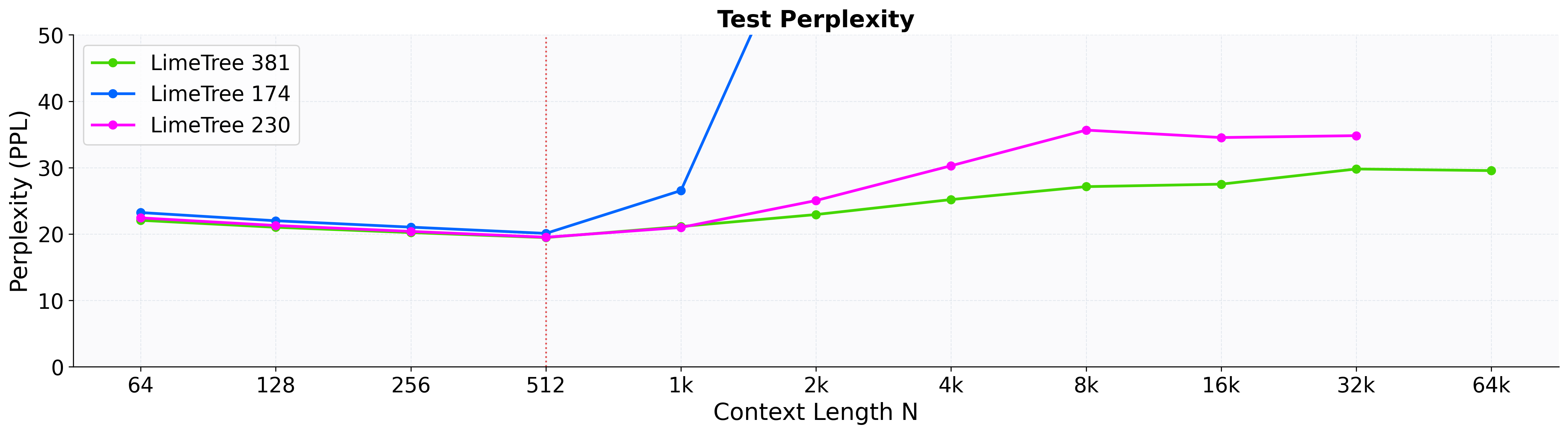}
    \caption{Comparison of LimeTree 174, 230 and 381. All three models show a similar PPL score at the training context length of 512, but LimeTree 381 demonstrates the strongst extrapolation capabilities up to 65536.}
    \label{fig:ctx_length_extrapolations} 
\end{figure}

To examine whether smaller, more efficient variants suffice, three LimeTree configurations ranging from $174\text{M}$ to $381\text{M}$ parameters were evaluated across context lengths. Up to the $512$-token training horizon, all three models perform comparably, reaching closely clustered test perplexities between $19.78$ and $20.40$. Once pushed beyond the training window, however, their behavior diverges sharply (Figure~\ref{fig:ctx_length_extrapolations}(a)). Whereas the $381\text{M}$-parameter model extrapolates smoothly to long horizons, the smallest $174\text{M}$ variant degrades rapidly, effectively collapsing once the context length is quadrupled. Sufficient parameter capacity is therefore essential for learning generalizable hierarchical tree representations, even when short-context performance appears saturated.

Placing these results into the broader context of word-level language modeling on WikiText-103 highlights how far linear-complexity architectures have progressed (Table~\ref{tab:wikitext103-benchmarks}). Early linear models such as Gated CNNs ($37.20$ PPL) and QRNNs ($33.00$ PPL) were fast to train but struggled to capture long-range dependencies, leaving a wide gap behind quadratic Transformers like Adaptive Input ($18.70$), Transformer-XL ($18.30$), and Compressive Transformer ($17.10$). More recently, state-space and linear attention models have narrowed this divide, with S4 reaching $20.95$ PPL, Hedgehog achieving $20.80$, and H3 and Mamba pushing down to $18.50$ and $17.90$. Despite being trained with a short $512$-token context window, the LimeTree family—reaching $20.40$ PPL at $174\text{M}$ parameters, $19.85$ at $174\text{M}$, and $19.78$ at $381\text{M}$—has lower perplexity than S4 and Hedgehog (but see the token-level vs word-level note) while approaching the perplexity regime of models trained on four to sixteen times longer contexts.

\begin{table}[htbp]
\centering
\caption{Closed-corpus token-level language modeling benchmark progression on WikiText-103.}
\label{tab:wikitext103-benchmarks}
\resizebox{\textwidth}{!}{%
\begin{tabular}{lccccc}
\toprule
\textbf{Model (Year)} & \textbf{Test PPL} & \textbf{Parameters} & \textbf{Context \& Complexity} & \textbf{GPUs} & \textbf{Time} \\
\midrule
Grave LSTM Cache (2016)~\cite{grave2017continuous} & 40.80 & 150M & 2000, Linear & 1 M40 & 96h \\
Gated CNN (2017)~\cite{dauphin2017language} & 37.20 & 229M & 200, Linear & 8 M40 & 168h \\
QRNN (2018)~\cite{merity2018analysis,bradbury2017quasi} & 33.00 & 151M & 140, Linear & 1 K80 & 36h \\
Adaptive Input (2018)~\cite{baevski2019adaptive} & 18.70 & 247M & 3072, Quadratic & 8 V100 & 63h \\
Transformer-XL (2019)~\cite{dai2019transformer} & 18.30 & 257M & 1600, Quadratic & 32 V100 & 60h \\
Compressive Transformer (2019)~\cite{rae2020compressive} & 17.10 & 277M & 3200, Quadratic & 32 TPUv3 & 48h \\
Adaptive Span (2019)~\cite{sukhbaatar2019adaptive} & 16.40 & 209M & 8192, Sub-quadratic & 16 V100 & 72h \\
Routing Transformer (2020)~\cite{roy2021efficient} & 15.80 & 230M & 4096, Sub-quadratic & 16 V100 & 84h \\
$k$NN-LM (2020)~\cite{khandelwal2020generalization} & 15.79 & 247M & 3072, Quadratic & 8 V100 & 71h \\
Shortformer (2021)~\cite{press2021shortformer} & 18.15 & 247M & 1024, Quadratic & 8 V100 & 34h \\
S4 State Space (2022)~\cite{gu2022efficiently} & 20.95 & 249M & 8192, Log-linear & 8 A100 & 36h \\
H3 State Space (2023)~\cite{fu2023hungry} & 18.50 & 248M & 2048, Log-linear & 8 A100 & 24h \\
Mamba (2023)~\cite{gu2024mamba} & 17.90 & 250M & 2048, Linear & 8 A100 & 18h \\
Hedgehog (2024) \cite{zhang2024hedgehogporcupineexpressive} & 20.80 & 150M(?) & Linear & - & - \\
\midrule
\textcolor{limegreen}{LimeTree 174} & \textcolor{limegreen}{20.40} & 174M & 512, Linear & \textcolor{limegreen}{4 Quadro RTX 8000} & \textcolor{limegreen}{8.46 h}\\
\textcolor{limegreen}{LimeTree 230} & \textcolor{limegreen}{19.85} & 230M  & 512, Linear & \textcolor{limegreen}{4 Quadro RTX 8000} & \textcolor{limegreen}{10.60 h}\\
\textcolor{limegreen}{LimeTree 381} & \textcolor{limegreen}{19.78} & 381M  & 512, Linear & \textcolor{limegreen}{4 Quadro RTX 8000} & \textcolor{limegreen}{15.56h}\\
\bottomrule
\end{tabular}
}
\end{table}

Equally striking in Table~\ref{tab:wikitext103-benchmarks} is the fraction of compute required to reach this performance, alongside the clear headroom for further gains. Competitive quadratic and sub-quadratic Transformers routinely required between $500$ and $1{,}920$ V100 GPU-hours (such as $32$ V100s for $60$ hours in Transformer-XL or $16$ V100s for $84$ hours in the Routing Transformer), while modern state-space baselines like S4, H3, and Mamba relied on $8$ datacenter A100 GPUs for $18$ to $36$ hours ($144\text{--}288$ A100 GPU-hours). By contrast, the LimeTree models were trained on just $4$ previous-generation Quadro RTX 8000 GPUs in $8.5$ to $15.6$ hours—a total budget of only $34$ to $62$ GPU-hours, well noted that the Quadro RTX 8000 is somewhat slower than the A100/H100. Furthermore, whereas every baseline below $19.0$ PPL in Table~\ref{tab:wikitext103-benchmarks} was trained with context lengths between $1{,}024$ and $8{,}192$ tokens, LimeTree was constrained to a $512$-token training horizon with a $256$-token causal tail. Scaling the training window to $2{,}048\text{--}4{,}096$ tokens and pairing the architecture with adaptive input embeddings provides a direct path to matching top attention baselines at a fraction of their training and inference cost.

\subsection{PG19 results}

\begin{figure}[htbp]
    \centering
    \includegraphics[width=1\linewidth]{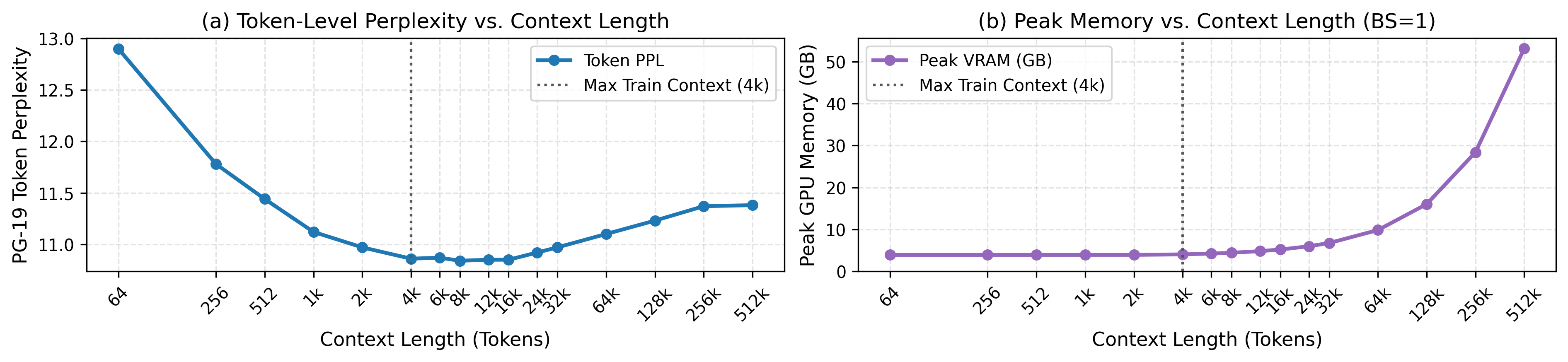}
    \caption{PG19 test evaluation with extrapolation.}
    \label{fig:limetree_pg19_extrapolation_curves}
\end{figure}

The second model with a slightly more complex architecture was trained on the dataset PG19 in 35.5h on a single Nvidia Quadro RTX 6000 Blackwell Server Edition. LimeTree achieved a token-level perplexity score of 10.84, and a corresponding word-level perplexity score of 37.49 when evaluated on the test set on the training context length of L=4096. Most remarkably LimeTree is able to hold a low perplexity score when evaluated over context lengths up to 512k, where it still has a PPL score of 11.38. Testing beyond the context length of 512k resulted in an out-of-memory exception on a single GPU. Table \ref{tab:gtformer_v2_scaling_step_44000} shows the detailed scores over all context lengths, together the the success rate in the key-value (KV) retrieval task. LimeTree also holds up very well in the KV task, with 100\% at most lengths up to 128k (95\% at 6k, 91\% at 12k), falling to 81\% at 256k and 50\% at 512k, while demonstrating linear memory increase (last two columns of Table \ref{tab:gtformer_v2_scaling_step_44000}). Figure \ref{fig:limetree_pg19_extrapolation_curves}(a) illustrates the evaluation of the perplexity score and  Figure \ref{fig:limetree_pg19_extrapolation_curves}(b) the increase in memory for the evaluation of LimeTree on the test set, both with logarithmic scaling on the horizontal axis.

\begin{table*}[htbp]
    \centering
    \scriptsize
    \setlength{\tabcolsep}{5.5pt}
    \caption{Multi-scale evaluation of LimeTree ($L=16, D=1536$, step 44000, $\text{BS}=1$) across context lengths up to 512k tokens. The horizontal rule marks the maximum training context length (4k tokens).}
    \label{tab:gtformer_v2_scaling_step_44000}
    \begin{tabular}{@{}l r r r r r r r r@{}}
        \toprule
        \textbf{Context} & \textbf{KV Exact-3 (\%)} & \textbf{KV Slot (\%)} & \textbf{NLL} & \textbf{Tok PPL} & \textbf{Word PPL} & \textbf{Throughput} & \textbf{Peak VRAM} & \textbf{Act. VRAM} \\
        \textbf{Length} & ($\uparrow$) & ($\uparrow$) & ($\downarrow$) & ($\downarrow$) & ($\downarrow$) & \textbf{(k tok/s)} & \textbf{(GB)} & \textbf{(GB)} \\
        \midrule
        64 & \textbf{100.00} & \textbf{100.00} & 2.5575 & 12.90 & 48.69 & 5.7 & 3.93 & +0.19 \\
        256 & \textbf{100.00} & \textbf{100.00} & 2.4661 & 11.78 & 42.37 & 9.1 & 3.93 & +0.20 \\
        512 & \textbf{100.00} & \textbf{100.00} & 2.4374 & 11.44 & 40.56 & 13.4 & 3.93 & +0.20 \\
        1k & \textbf{100.00} & \textbf{100.00} & 2.4092 & 11.12 & 38.86 & 22.4 & 3.93 & +0.20 \\
        2k & \textbf{100.00} & \textbf{100.00} & 2.3955 & 10.97 & 38.06 & 38.9 & 3.93 & +0.20 \\
        4k & \textbf{100.00} & \textbf{100.00} & 2.3854 & 10.86 & 37.49 & 69.5 & 4.05 & +0.31 \\
        \midrule
        6k & 95.31 & 98.44 & 2.3860 & 10.87 & 37.52 & 100.2 & 4.24 & +0.51 \\
        8k & \textbf{100.00} & \textbf{100.00} & \textbf{2.3832} & \textbf{10.84} & \textbf{37.36} & 136.3 & 4.43 & +0.70 \\
        12k & 90.62 & 96.88 & 2.3841 & 10.85 & 37.41 & 197.3 & 4.82 & +1.09 \\
        16k & \textbf{100.00} & \textbf{100.00} & 2.3843 & 10.85 & 37.42 & 265.2 & 5.21 & +1.48 \\
        24k & \textbf{100.00} & \textbf{100.00} & 2.3908 & 10.92 & 37.79 & 355.2 & 5.98 & +2.25 \\
        32k & \textbf{100.00} & \textbf{100.00} & 2.3956 & 10.97 & 38.07 & 407.7 & 6.75 & +3.02 \\
        64k & \textbf{100.00} & \textbf{100.00} & 2.4071 & 11.10 & 38.74 & 510.2 & 9.85 & +6.12 \\
        128k & \textbf{100.00} & \textbf{100.00} & 2.4185 & 11.23 & 39.42 & 572.4 & 16.04 & +12.30 \\
        256k & 81.25 & 93.75 & 2.4311 & 11.37 & 40.18 & 589.3 & 28.41 & +24.68 \\
        512k & 50.00 & 81.25 & 2.4319 & 11.38 & 40.23 & 598.8 & 53.17 & +49.44 \\
        \bottomrule
    \end{tabular}
\end{table*}

A detailed literature search for related benchmarking results was undertaken and Table \ref{tab:pg19_models} was compiled at the best effort. Benchmarking results in the literature are often difficult to compare, since evaluation techniques differ. Models throughout the literature were naturally trained on different computational setups, making also a training time comparison challenging. However, as Table \ref{tab:pg19_models} demonstrates, LimeTree scores very well for a minimally optimized model, with a token-level perplexity score of 10.84, which is right in the range of other models evaluated on a token-level. The word-level perplexity score of 37.36 is slightly higher than the baseline of the Transformer-XL, but given the early days of LimeTree, it already shows great potential. Most significantly, LimeTree reached these scores with markedly fewer computational resources during training; although the setups are not directly comparable, even a rough comparison indicates a substantial difference.

\begin{table*}[htbp]
\centering
\caption{Comparison of sequence modeling architectures evaluated on the PG-19 long-range book benchmark, highlighting sub-quadratic models alongside canonical quadratic/cached baselines. Complexity is reported per layer with respect to sequence length $N$, segment/window size $W$, latent/state dimension $d$ (or latent count $L$), and external memory size $M$. NR denotes metrics not reported by the original authors.}
\label{tab:pg19_models}
\resizebox{\textwidth}{!}{%
\begin{threeparttable}
\begin{tabular}{@{}llrrllllr@{}}
\toprule
\textbf{Model} & \textbf{Citation} & \textbf{Params} & \textbf{Complexity} & \textbf{Train Ctx} & \textbf{Eval Ctx} & \textbf{Compute Infrastructure} & \textbf{Train Time / Steps} & \textbf{Test PPL $\downarrow$} \\
\midrule
\multicolumn{9}{@{}l}{\textit{\textbf{Panel A: From-Scratch PG-19 Training --- Official Word-Level Perplexity} (6.97M test words)}} \\
\midrule
Transformer-XL (36L) & \cite{rae2020compressive} & ${\sim}529$M & $\mathcal{O}(W^2 + W M)$ & $512{+}1024$ & $1536$ & $64\times$ TPUv3 cores & ${\sim}24$h (NR steps) & 36.30 \\
Compressive Transformer (36L) & \cite{rae2020compressive} & ${\sim}529$M & $\mathcal{O}(W^2 + W(M{+}M_c))$ & $512{+}2048$ & $2560$ & $64\times$ TPUv3 cores & $>12$--$24$h & 33.60 \\
Routing Transformer (22L) & \cite{roy2021efficient} & 490M & $\mathcal{O}(N^{1.5} d)$ & 8,192 & 8,192 & $128\times$ TPUv3 cores & ${\sim}250$k steps & 33.20 \\
Perceiver AR (60L) & \cite{hawthorne2022general} & 975M & $\mathcal{O}(N L d + L^2 d)$ & 2,048 & 2,048 / 4,096 & TPUv3 Pod & $200$k steps (420B tok) & 28.90 / 29.00 \\
Block-Recurrent Trans. (24L) & \cite{hutchins2022block} & 650M & $\mathcal{O}(N W d)$ & 4,096 ($W{=}512$) & Full book & $32\times$ TPUv4 cores & ${\sim}48$h ($500$k steps) & 28.46 \\
Block-Recurrent Trans. (24L) & \cite{hutchins2022block} & 1.3B & $\mathcal{O}(N W d)$ & 4,096 ($W{=}512$) & Full book & $32\times$ TPUv4 cores & $>48$h ($500$k steps) & 26.50 \\
\textcolor{limegreen}{\textbf{LimeTree 997}} & \textcolor{limegreen}{\textbf{Current Work (2026)}} & \textbf{997M (996,558,145)}\tnote{a} & $\mathcal{O}(N W_c d + T^2 d)$ & \textbf{256--4,096} & \textbf{up to 512k} & 1× NVIDIA RTX PRO 6000 BSE & \textbf{$44$k steps/35.5h} & 37.36 \\
\midrule
\multicolumn{9}{@{}l}{\textit{\textbf{Panel B: From-Scratch PG-19 Training --- Subword / Token-Level Perplexity} (32k SentencePiece or 50k GPT-2 BPE)}} \\
\midrule
Performer (FAVOR+, 36L) & \cite{choromanski2021rethinking} & ${\sim}200$M & $\mathcal{O}(N m d)$ & 4,096 / 8,192 & 4,096 / 8,192 & TPUv3 / V100-16GB & $125$k--$250$k steps & Curves only\tnote{b} \\
Transformer-XL (12L) & \cite{munkhdalai2024leave} & ${\sim}183$M & $\mathcal{O}(W^2 + W M)$ & 2,048 ($M{=}2048$) & 4,096 & TPUv4 / TPUv5e & NR & 11.88 \\
Memorizing Transformer (12L) & \cite{wu2022memorizing} & ${\sim}183$M & $\mathcal{O}(N W d + N \log M)$ & 2,048 ($M{=}65$k) & 65,536 & $32\times$ TPUv3 / TPUv4 & $500$k steps & 11.37 \\
Diagonal State Spaces (DSS) & \cite{gupta2022diagonal,mehta2023long} & 209M & $\mathcal{O}(N \log N)$ & 4,096 & 4,096 / 16,384 & TPUv4 ($1.8$ steps/s) & ${\sim}19.3$h ($125$k steps) & 13.53 / 692.1 \\
Gated State Spaces (GSS, 16L) & \cite{mehta2023long} & 192M & $\mathcal{O}(N \log N)$ & 4,096 & 4,096 / 65,536 & TPUv4 ($5.3$ steps/s) & ${\sim}6.5$h ($125$k steps) & 12.84 / 12.47 \\
Block-State Transformer (BST) & \cite{fathi2023block} & ${\sim}200$M & $\mathcal{O}(N \log N + N W d)$ & 4,096 & 4,096+ & TPUv4 cores & Matched TPUv4-h & $<$GSS-Hybrid\tnote{c} \\
Recurrent Memory Trans. (RMT) & \cite{bulatov2022recurrent,munkhdalai2024leave} & ${\sim}183$M & $\mathcal{O}(N W d)$ & 8,192 ($W{=}2048$) & 32,768 & TPUv4 / TPUv5e & NR & 13.27 \\
Infini-Transformer (Linear) & \cite{munkhdalai2024leave} & ${\sim}183$M & $\mathcal{O}(N W d + N d^2)$ & 32,768 ($W{=}2048$) & 32,768+ & TPUv4 / TPUv5e & NR & 9.65 \\
Infini-Transformer (+Delta) & \cite{munkhdalai2024leave} & ${\sim}183$M & $\mathcal{O}(N W d + N d^2)$ & 32,768 ($W{=}2048$) & 32,768+ & TPUv4 / TPUv5e & NR & 9.67 \\
Landmark Attention (12L) & \cite{mohtashami2023landmark} & 124M & $\mathcal{O}(N B d + N \frac{N}{B} d)$ & 512 & 4,096--32,768 & NVIDIA A100 GPUs & ${\sim}15$B tokens & 16.23 \\
MemMamba (SSM + Mem Gate) & \cite{wang2025memmambarethinkingmemorypatterns} & ${\sim}130$M & $\mathcal{O}(N d^2)$ & 4,096--8,192 & 60,000 & NVIDIA A100 GPUs & NR & 17.35 \\
MS-GLA (Multi-Scale GLA) & \cite{dev2026msglamultiscalegatedlinear} & 340M & $\mathcal{O}(N d^2)$ & 2,048 & 2,048--30,720 & NVIDIA A100 / H100 & Chinchilla-matched & $-9.5\%$ vs.\ GLA \\
\textcolor{limegreen}{LimeTree 997} & \textcolor{limegreen}{\textbf{Current Work (2026)}} & 997M & & 512-4096 & up to 512k & 1× NVIDIA RTX PRO 6000 BSE & 35.5h & 10.84 \\
\midrule
\multicolumn{9}{@{}l}{\textit{\textbf{Panel C: External Multi-Corpus Pretraining (RedPajama / 2T Web Tokens) --- Zero-Shot / Adapted Token PPL}}} \\
\midrule
LongLoRA ($\text{S}^2$-Attn, Llama-2) & \cite{chen2024longlora} & 7B & $\mathcal{O}(N W_g d)$ (FT) & 32,768 (FT) & 32,768 & $8\times$ A100-80GB & $1$k FT steps & 7.96 (val) \\
Megalodon (CEMA + NormAttn) & \cite{ma2024megalodon} & 7B & $\mathcal{O}(N W_c d)$ & 4,096--32,768 & Up to 2M & Multi-node A100/H100 & 2T tokens & ${\sim}6.8$--$7.2$ \\
\bottomrule
\end{tabular}
\end{threeparttable}%
}
\end{table*}

These initial empirical findings provide a strong proof of concept for the proposed approach. Although LimeTree does not yet surpass leading baselines on the benchmark leaderboards, its competitive performance demonstrates clear architectural promise. These encouraging outcomes justify further refining the core architecture, optimizing the training pipeline, and evaluating LimeTree at scale across a broader suite of benchmark datasets.

The present study serves as an initial proof of concept, and several methodological considerations frame the interpretation of these results. Because the external benchmarks in Tables 3 and 5 rely on published figures rather than compute-matched training runs, cross-model variations in tokenization, perplexity granularity (token-level versus word-level), parameter count, context length, and hardware render the training-cost comparisons indicative rather than directly equivalent. Furthermore, while the architecture achieves linear encoder scaling with sequence length, it remains a hybrid rather than strictly attention-free design, pairing the encoder with a 256-token causal Transformer decoder that cross-attends to a bounded memory of at most 193 slots; isolating the specific contribution of the tree memory from the local convolution, chunked attention, and decoder components awaits systematic ablation. On the synthetic recall benchmark, retrieval remained exact across most lengths up to 128k tokens before moderating to 81\% at 256k and 50\% at 512k, though the compact setup (three keys and four values) Primarily tests long-range retention rather than associative memory capacity, and length extrapolation was not observed in the smallest WikiText-103 model. Extending this evaluation—currently based on single runs per configuration—to budget-matched Transformer and state-space baselines, multi-seed ablations, higher-capacity recall tasks, and downstream or streaming benchmarks represents a natural next step for future work.

\section{Conclusion}

This work introduces LimeTree as a scalable alternative to standard Transformer architectures, competing directly against a diverse range of modern sequence-modeling variants. Across baseline comparisons, LimeTree immediately achieves solid mid-table benchmark performance—an encouraging foundation for a novel design that has not yet undergone the years of heavy hyperparameter and hardware optimization afforded to standard Transformers. Crucially, LimeTree exhibits characteristics that make it exceptionally compelling for next-generation Large Language Models (LLMs). Its memory footprint and computational cost scale strictly linearly with sequence length, breaking the quadratic attention bottleneck to enable lightweight yet high-capacity foundation models. This efficiency is paired with remarkable zero-shot length extrapolation: despite training on sequences of at most $4,096$ tokens, LimeTree suffers about 5\% higher perplexity (10.84 $\rightarrow$ 11.38) at 524,288 tokens, with key-value recall down to 50\% ($128\times$ extrapolation). By combining linear complexity with stable long-range reasoning, LimeTree offers a viable path toward LLMs capable of processing massive contexts without prohibitive training or inference costs.

\section*{Acknowledgements}

The author would like to thank her colleagues at Middlesex University London for on-going excellent support with the computing infrastructure and for kindly tolerating extensive use of GPUs. 

\section*{Declaration of AI usage}

Generative AI tools (Google Gemini) were used throughout this project to assist with literature search, code development and debugging, and drafting and refining the manuscript text as well as generating the image in the title. All generated code, experimental implementations, and written content were critically reviewed, verified, and validated by the author, who takes full responsibility for the integrity and accuracy of this work.

\bibliographystyle{plain}  
\bibliography{bibliography}

\end{document}